\documentclass{article} % For LaTeX2e
\usepackage{times}
\usepackage{amsmath,amsfonts,bm}

\def\eqref#1{equation~\ref{#1}}
\def\1{\bm{1}}

\DeclareMathAlphabet{\mathsfit}{\encodingdefault}{\sfdefault}{m}{sl}
\SetMathAlphabet{\mathsfit}{bold}{\encodingdefault}{\sfdefault}{bx}{n}

\usepackage{hyperref}
\usepackage{url}
\usepackage[utf8]{inputenc} % allow utf-8 input
\usepackage[T1]{fontenc}    % use 8-bit T1 fonts
\usepackage{hyperref}       % hyperlinks
\usepackage{url}            % simple URL typesetting
\usepackage{booktabs}       % professional-quality tables
\usepackage{amsfonts}       % blackboard math symbols
\usepackage{nicefrac}       % compact symbols for 1/2, etc.
\usepackage{microtype}      % microtypography
\usepackage{xcolor}         % colors
\usepackage{tikz}
\usepackage{graphicx}
\usepackage{amsthm}
\usepackage{algorithm}
\usepackage{algorithmic}
\usepackage{algorithm}
\usepackage{natbib}
\usepackage{multicol}
\usepackage{multirow}
\usepackage{bm}
\usepackage{algorithmic}
\usepackage{amsmath}
\newtheorem{assumption}{Assumption}
\usepackage{cleveref}
\usepackage{caption}
\newtheorem{theorem}{Theorem}

\usepackage{times}
\usepackage{soul}
\usepackage{url}
\title{Detecting Scarce and Sparse Anomalous: Solving Dual Imbalance in Multi-Instance Learning}
\author{Lin-Han Jia$^{1}$, Lan-Zhe Guo$^{1,3}$, Zhi Zhou$^{1,2}$, \\Si-Yu Han$^{1,2}$, Zi-Wen Li$^{4}$, Yu-Feng Li$^{1,2}$\\
National Key Laboratory for Novel Software Technology$^{1}$\\
School of Artificial Intelligence$^{2}$\\
School of Intelligence Science and Technology$^{3}$\\
Nanjing University, Nanjing 210023, China\\
Didi Chuxing, China$^{4}$\\
\texttt{\{jialh,guolz,zhouz,liyf\}@lamda.nju.edu.cn}\\
\texttt{liziwen@didiglobal.com}
}
\begin{document}

\maketitle
\footnotetext{Corresponding Author: Yu-Feng Li and Lan-Zhe Guo.}
\begin{abstract}
In real-world applications, it is highly challenging to detect anomalous samples with extremely sparse anomalies, as they are highly similar to and thus easily confused with normal samples. Moreover, the number of anomalous samples is inherently scarce. This results in a dual imbalance Multi-Instance Learning (MIL) problem, manifesting at both the macro and micro levels. To address this "needle-in-a-haystack problem", we find that MIL problem can be reformulated as a fine-grained PU learning problem. This allows us to address the imbalance issue in an unbiased manner using micro-level balancing mechanisms. To this end, we propose a novel framework, Balanced Fine-Grained Positive-Unlabeled (BFGPU)-based on rigorous theoretical foundations. Extensive experiments on both synthetic and real-world datasets demonstrate the effectiveness of BFGPU. 
% In such cases, both macro-level and micro-level detection encounter significant difficulties. The former struggles because normal and anomalous samples are highly similar and hard to distinguish at the macro level, while the latter struggles because of the lack of labels at the micro level. 
% In MIL, micro-level labels are inferred from macro-level labels, which can lead to severe bias. Moreover, the more imbalanced the distribution between normal and anomalous samples, the more pronounced these limitations become. 
% In this study, we observe that the MIL problem can be elegantly transformed into a fine-grained Positive-Unlabeled (PU) learning problem. This transformation allows us to address the imbalance issue in an unbiased manner using a micro-level balancing mechanism.
% To this end, we propose a novel framework-Balanced Fine-Grained Positive-Unlabeled (BFGPU)-based on rigorous theoretical foundations to address the challenges above. Extensive experiments on both public and real-world datasets demonstrate the effectiveness of BFGPU, which outperforms existing methods, even in extreme scenarios where both macro and micro-level distributions are highly imbalanced. 
% The code is open-sourced at https://github.com/BFGPU/BFGPU.
\end{abstract}
\section{Introduction}
\label{introduction}
In real-world applications, there is a strong demand for effective detection of anomalous samples, such as in quality inspection, risk control, and fault diagnosis ~\citep{gornitz2013toward,pang2023deep,ye2023uadb}. Ideally, anomalous samples exhibit significant and easily distinguishable differences from normal samples, making them readily separable based on features or representations. However, in more realistic scenarios, the differences between anomalous and normal samples are often subtle because anomalies may manifest as sparse local information. As a result, anomalous samples are highly similar to normal ones at a macro level. Moreover, the number of anomalous samples is inherently scarce in reality, making them more difficult to identify.

\begin{figure*}[htbp]
% \vskip 0.2in
\begin{center}
\centerline{\includegraphics[scale=0.4]{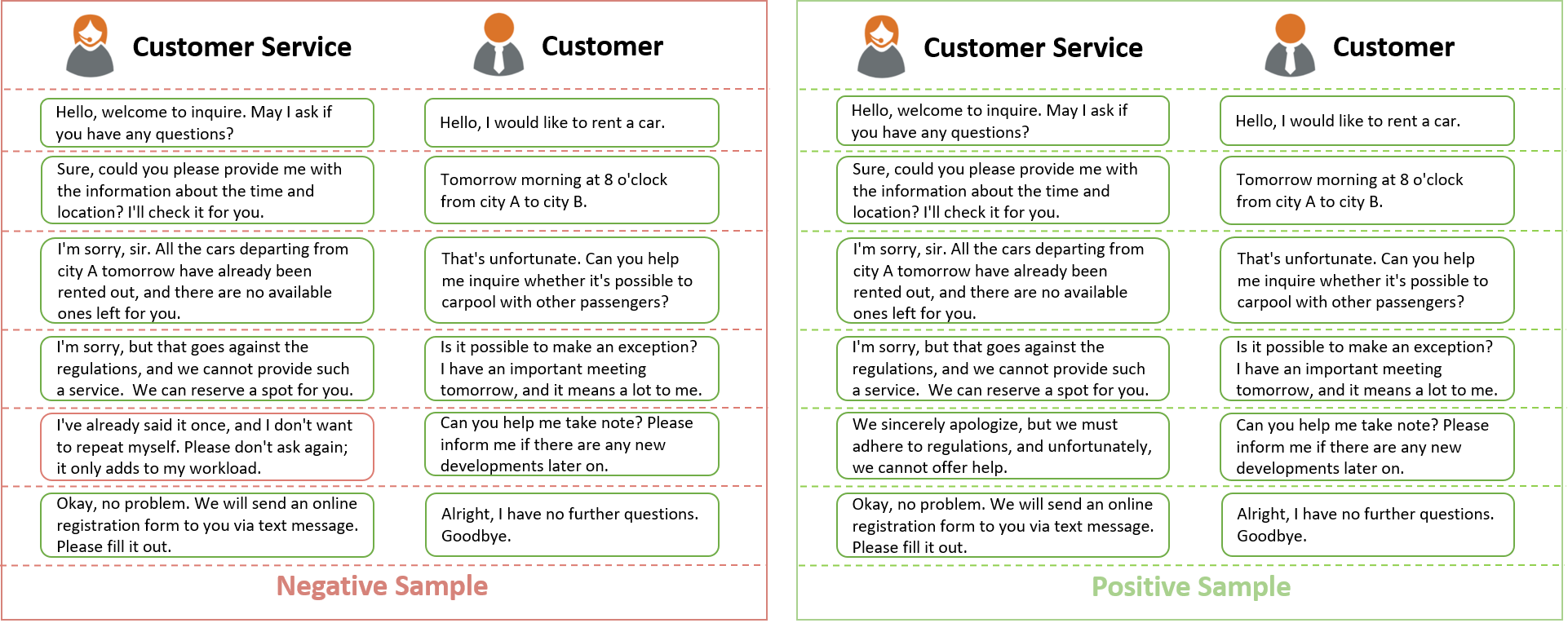}}
\caption{This example illustrates the difficulty in distinguishing between normal and anomalous macro samples due to the low proportion of anomalous information within anomalous samples.}
\label{Data}
\end{center}
\vskip -0.4in
\end{figure*}

On the one hand, this problem has significant practical relevance. For instance, in cancer cell detection, diseased samples are inherently rare, and within these samples, the pathological regions often occupy only a minuscule portion. This makes early detection before the cancer metastasizes exceptionally challenging. On the other hand, the rise of large language models (LLMs) has further amplified the importance of this detection problem. During the Reinforcement Learning from Human Feedback (RLHF) phase, aligning LLMs with human values requires training a separate detector as a reward model \citep{bai2022training}. This model is trained on human feedback to identify non-compliant sections in the LLM's outputs. However, these non-compliant contents are both scarce and sparse. Compounding this is the high cost of human annotation, which typically only provides coarse-grained labels. These factors make training an accurate reward model extremely difficult, thereby further highlighting the critical nature of this detection challenge. An example is provided in \cref{Data}, a real-world customer service quality inspection task, the goal is to detect instances of impoliteness within extended conversations between customer service and customer. Since the customer service is well-trained, mistakes—if any—typically occur in only a subtle utterance.
% 一方面这一问题本身有很强的现实基础，例如在癌细胞检测的任务中，患病样本本身非常稀少，且在患病样本中发生病变的部分也仅仅占很小比例，难以在癌细胞大范围扩散前被及时发现。另一方面随着大语言模型的火热，由于其在RLHF阶段需要实现与人类价值观的对齐，需要用人类反馈数据训练额外的检测器作为奖励模型用于检测模型输出不符合规范的部分，但由于模型输出的不合规内容稀疏且细微，人类标注昂贵且只有针对整段回答的宏观标注，导致奖励模型极难训练，进一步这一检测问题的重要性被进一步放大。图一给了一个例子，

Given the difficulty of distinguishing samples at the macro level, a promising approach is to seek solutions at the micro level—by decomposing each macro-level sample into multiple finer-grained components and performing discrimination at this finer resolution. However, there are no precise labels available at the micro level, so learning must rely solely on macro-level supervision. Multiple Instance Learning (MIL) is a paradigm of weakly supervised learning designed for scenarios with imprecise labels ~\citep{waqas2024exploring,zhou2018brief,gao2022discrepant}. It structures data into two hierarchical levels: the macro level (referred to as bags) and the micro level (referred to as instances), and utilizes only macro-level labels for training. To address the lack of micro-level supervision, existing MIL methods often heuristically assign bag labels to all instances within the bag ~\citep{ilse2018attention,angelidis2018multiple,pang2023deep}. However, this strategy lacks theoretical grounding and can introduce substantial bias. In terms of data imbalance, the micro-level imbalance—known in MIL literature as the low witness rate problem—has been recognized but remains unresolved ~\citep{carbonneau2018multiple,zhang2022dtfd}. Meanwhile,macro-level imbalance has been largely overlooked and presents an even greater challenge.
In this paper, instead of directly assigning bag labels to instances, we treat all instances from normal bags as positive instances and all instances from anomalous bags as unlabeled instances. This reformulates the MIL problem as a fine-grained Positive-Unlabeled (PU) learning task. PU learning is a learning paradigm that trains models using only positive and unlabeled data ~\citep{bekker2020learning}. 
% Its core assumption is that both labeled and unlabeled positive instances are drawn from the same underlying distribution, which enables estimation of the distribution of negative instances within the unlabeled set and thus facilitates the separation of positive and negative instances—thereby alleviating the label bias challenge in MIL ~\citep{du2015convex,kiryo2017positive}. 
Building upon this reformulation, we further derive a PU learning loss function based on rigorous theoretical analysis, which simultaneously addresses the dual imbalance challenge in MIL.
% imbalance at both the macro and micro levels—thereby alleviating 
% 为了得到更加适用于解决MIL问题的PU学习算法，我们进一步考虑在微观层面的学习中利用宏观的信息，仅对自信度高的无标注严格不能添加伪标注，并通过动态调整阈值的策略保持模型对正常样本与异常样本分类的平衡性，最终使模型可以直接在微观层面提升宏观分类性能，形成了用于解决MIL问题的PU学习算法BFGPU。

To develop a PU learning algorithm better suited for addressing MIL problems, we further incorporate macro-level information into the micro-level learning process. Specifically, we restrict the assignment of pseudo-labels to only high-confidence unlabeled instances, and dynamically adjust the confidence threshold to maintain balanced prediction. This design enables the model to improve macro-level performance directly through micro-level learning. The resulting algorithm, termed Balanced Fine-Grained Positive-Unlabeled (BFGPU), is a PU learning method tailored for solving MIL problems.
\begin{figure}[htbp]
\vskip -0.2in
\begin{center}
\centerline{\includegraphics[scale=0.33]{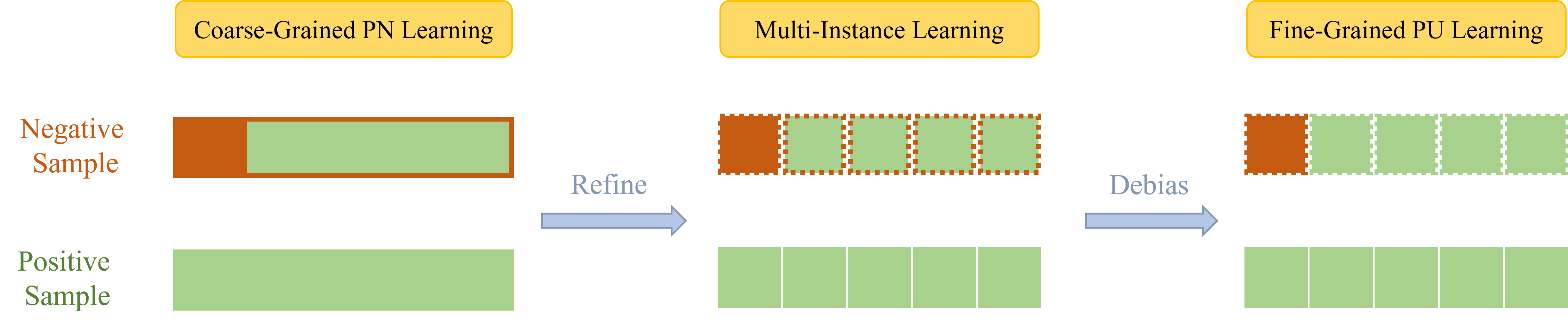}}
\caption{This figure illustrates the different solution paradigms for the detection problem in this paper: coarse-grained positive-negative learning, MIL, and fine-grained positive-unlabeled learning.}
\label{Paradiam}
\end{center}
\vskip -0.35in
\end{figure}
In the text and image modalities, we conducted comprehensive comparisons with existing supervised learning, anomaly detection, MIL, and PU learning methods based on customer service quality inspection (CSQI) and invasive ductal carcinoma (IDC) tasks, respectively. Furthermore, to validate the robustness of our algorithm under varying macro and micro imbalance ratios, we synthesized numerous datasets with different imbalance ratios based on four fundamental sentiment analysis datasets for additional comparative experiments. We also compared its performance with popular LLMs in addressing the "needle in a haystack" problem~\citep{wang2024needle,kuratov2024babilong}. Finally, through ablation studies and parameter sensitivity analysis, we further enriched the discussion on the algorithm's robustness.
% We conducted extensive experiments to evaluate the performance of BFGPU across various datasets. , including 2 datasets with a maximum imbalance ratio of 5 and 2 datasets with a maximum imbalance ratio of 10. Additionally, we tested the algorithm on a real-world customer service quality inspection dataset. These experiments demonstrated the effectiveness of BFGPU in both diverse datasets and practical applications. Furthermore, we explored scenarios with both macro and micro imbalances, and even in extreme cases where anomalous information is vastly outnumbered, our algorithm consistently performed well.

% 我们进行了大量实验验证了我们提出的方案的性能，包括3组最高不平衡比例为5和2组最高不平衡比例为10的数据集，以及在真实场景中遇到的客服质量检测数据集，这些实验都证明了算法在广泛数据集和实际应用中的有效性。并且我们进一步在宏观和微观上都不平衡的问题上进行了实验，发现即使在这种负类信息远远小于正类信息的极端场景下我们的算法依然表现良好。
% 在这篇文章中，我们的贡献主要分为三个部分：1. 我们通过连接宏观和微观的学习目标，为解决高度相似文本的分类提供了新的学习范式；2. 我们分析了宏观PN分类和微观不平衡PU分类各自的误差和适用范围，为在二者间进行选择提供了依据；3. 我们为解决上述问题提出了新的算法，在实验中取得了优异的效果。
% \textbf{Our Contributions.} In this paper, our contributions are primarily divided into three parts. Firstly, We introduce a new learning paradigm by connecting macro and micro learning objectives, providing a novel approach to address the classification of highly similar texts. Secondly, We analyze the errors and applicability of macro PN classification and micro imbalanced PU classification, providing a basis for choosing between the two. Thirdly, We propose a new algorithm to address the aforementioned issues, achieving outstanding results in experimental evaluations.
\textbf{Our Contributions.} 1. We formalize the challenging detection problem in practical applications and propose a dual imbalanced MIL problem. 2. We propose a novel perspective that reformulates the MIL problem as a balanced PU learning task, which addresses the issues of bias and imbalance. 3. We incorporate macro information into the learning process, leading to BFGPU, an algorithm that enables balanced and unbiased macro performance by training at the micro level. 4. We provide a theoretical analysis demonstrating the advantages of BFGPU over macro-level learning and conventional MIL approaches. 5. Extensive experiments validate the superior performance of BFGPU.
\section{Related Works}
% \subsection{Positive Unlabeled Learning}
\textbf{PU learning} can utilize positive data and unlabeled data to train a classifier that distinguishes between positive and negative data ~\citep{bekker2020learning}. 
% The core idea is that both labeled and unlabeled positive samples are sampled from the same distribution. 
% It is reasonable to use the loss of unlabeled data, proportionally subtracted from the loss of positive data, to estimate the loss of negative data. 
Most PU learning algorithms aim for the model to be unbiased in terms of accuracy, and there are many effective algorithms currently available ~\citep{du2015convex,kiryo2017positive,chen2020self,kato2018learning,shi2018positive,sansone2018efficient,hou2018generative,hsieh2019classification,hsieh2015pu,wang2024pue}. Some PU learning algorithms address the issue of imbalance in the positive-negative ratio within unlabeled data and strive for unbiased average accuracy (AvgAcc) ~\citep{su2021positive}. Additionally, theoretical research on PU learning has demonstrated its superiority
% over PN learning under certain conditions 
~\citep{niu2016theoretical}.

% \subsection{Multi-Instance Learning}
% Weakly supervised learning is a paradigm designed to address situations where label information is scarce ~\citep{zhou2018brief}.
% It mainly includes incomplete supervised learning, where labels are missing ~\citep{van2020survey}; inaccurate supervised learning, where label errors may occur ~\citep{song2022learning}; and inexact supervised learning, also known as 
\textbf{Multi-instance learning (MIL)} is a form of weakly supervised learning ~\citep{zhou2018brief} characterized by inexact supervision, where labels are not precise ~\citep{carbonneau2018multiple}. In MIL, a single label is assigned to a bag of instances, addressing the problem we are discussing. In the past, MIL approaches typically involved pooling or attention mechanisms to fuse embeddings or setting up scoring functions ~\citep{ilse2018attention,feng2017deep,pinheiro2015image,perini2023learning,pang2023deep,abati2019latent}. In our case, we aim to transform this problem into a direct micro PU learning problem to achieve more accurate solutions and eliminate redundant information.

\textbf{Anomaly Detection (AD)} the process of detecting data instances that significantly deviate from the majority of data instances ~\citep{pang2021deep}.
% In the field of AD, there have been many classical algorithms ~\citep{manevitz2001one,liu2008isolation}, and 
% With the development of deep learning, relying on excellent representation capabilities, some deep AD algorithms have achieved success on complex forms of data ~\citep{wu2019deep,perera2019ocgan,zheng2019one,golan2018deep,abati2019latent}. 
% Deep SAD can simultaneously use labeled and unlabeled data to learn an anomaly detector ~\citep{ruff2019deep}. This method utilizes representations obtained from neural networks. It initially fixes a center for normal samples and continuously pulls normal and unlabeled samples closer to the center while pushing anomalous samples away from the center. 
Historically, AD methods have been categorized into supervised AD ~\citep{gornitz2013toward}, weakly supervised AD ~\citep{ruff2019deep,pang2023deep, perini2023learning}, and unsupervised AD ~\citep{ruff2018deep,ye2023uadb}. 
% Deep SAD fixes a center in the embedding space for normal samples and continuously pulls normal and unlabeled samples closer to the center while pushing anomalous samples away.
However, there hasn't been an effective AD algorithm tailored for the form where anomalous are both scarce and sparse. 
% Therefore, existing methods cannot be used to address the proposed problem, as macro AD and macro PN classification are similarly ineffective due to high similarity.
\section{Derivation of Unbiased and Balanced PU Learning}
In classical supervised learning problems, we often deal with positive-negative (PN) learning. Assume there is an underlying distribution $p(x, y)$, where $x\in\mathbb{R}^d$ is the input, and $y \in\{-1, +1\}$ is the output. Data of size $n_+$ are sampled from $p(x|y = +1)$, and data of size $n_-$ are sampled from $p(x|y=-1)$. We let $g: \mathbb{R}^d \rightarrow \mathbb{R}^2$ be a decision function from a function space $\mathcal{G}$, where $g_{-1}$ and $g_{+1}$ are the probabilities of the sample being negaftive and positive, respectively. We also let $\mathcal{L}: \mathbb{R}^d \times \{-1, +1\} \rightarrow \mathbb{R}$ be a loss function.
% The based loss function is the $0-1$ loss defined as 
% \begin{align}
% \mathcal{L}_{0-1}[g(x), y]=\mathbb{I}[\frac{1}{2}(y+1)==\mathbb{I}[g_{+1}(x)\ge 0.5]] 
% \end{align}
% where $\mathbb{I}$ is the indicator function. 
The goal of PN learning is to use P data and N data to learn a classifier, denoted as $f: \mathbb{R}^d \rightarrow \{-1, +1\}$ which is based on the decision function $g$. 

However, in some scenarios, we may only have labels for one class, either only positive class labels known as PU learning or only negative class labels known as Negative-Unlabeled (NU) learning which is equivalent. In PU learning, data of size $n_+$  is sampled from $p(x|y = +1)$, and data of size $n_u$ is sampled from $p(x)$. The goal is also to learn a classifier $f: \mathbb{R}^d \rightarrow \{-1, +1\}$ which is the same as PN learning. We let $\pi = p(y = +1)$ represent the class prior of the positive data. The learning objective $R_{pn}(g)=E_{p(x,y)}[\mathcal{L}[(g(x),y]]$ of PN learning can be decomposed as
% \begin{align}
% R_{pn}(g)&=(1-\pi)E_{p(x|y=-1)}[\mathcal{L}[g(x),-1]]\nonumber\\&+\pi E_{p(x|y=+1)}[\mathcal{L}[g(x),+1]].
% \end{align} 
\begin{align}
R_{pn}(g)=(1-\pi)E_{p(x|y=-1)}[\mathcal{L}[g(x),-1]]+\pi E_{p(x|y=+1)}[\mathcal{L}[g(x),+1]].
\end{align} 
Du Plessis et al.~\citep{du2015convex} derived an unbiased PU (uPU) learning objective. Kiryo et al.~\citep{kiryo2017positive} propose non-negative PU (nnPU), avoiding the situation in uPU where the non-negative part of the loss is negative. In them, due to the absence of labeled negative samples for PU learning, the term $E_{p(x|y=-1)}[\mathcal{L}[g(x),-1]]$ cannot be directly estimated. However, since
% \begin{align}
% E_{p(x)}[\mathcal{L}[g(x),-1]]&=(1-\pi)E_{p(x|y=-1)}[\mathcal{L}[g(x),-1]]\nonumber\\&+\pi E_{p(x|y=+1)}[\mathcal{L}[g(x),-1]], 
% \end{align}
\begin{align}
E_{p(x)}[\mathcal{L}[g(x),-1]]=(1-\pi)E_{p(x|y=-1)}[\mathcal{L}[g(x),-1]]+\pi E_{p(x|y=+1)}[\mathcal{L}[g(x),-1]], 
\end{align}
where $\pi$ is the proportion of positive samples, $E_{p(x|y=-1)}[\mathcal{L}[g(x),-1]]$ can be indirectly estimated by leveraging the unlabeled data to estimate $E_{p(x)}[\mathcal{L}[g(x),-1)]]$ and using positive data to estimate $E_{p(x|y=+1)}[\mathcal{L}[g(x),-1]]$. So the empirical unbiased PU learning loss can be estimated using the current model's prediction $\hat{g}$ as
% \begin{align}
% \hat{R}_{upu}(g)&=\frac{\pi}{n_p} \sum_{x_i\in P}\mathcal{L}[g(x_i),+1+\frac{1}{n_u}\sum_{x_i\in U}\mathcal{L}[g(x_i),-1]]\nonumber\\&-\frac{\pi}{n_p}\sum_{x_i\in P}\mathcal{L}[g(x_i),-1].
% \end{align}
\begin{align}
\hat{R}_{upu}(\hat{g})=\frac{\pi}{n_p} \sum_{x_i\in P}\mathcal{L}[\hat{g}(x_i),+1]+\frac{1}{n_u}\sum_{x_i\in U}\mathcal{L}[\hat{g}(x_i),-1]-\frac{\pi}{n_p}\sum_{x_i\in P}\mathcal{L}[\hat{g}(x_i),-1].
\end{align}
% Kiryo et al.~\citep{kiryo2017positive} proposed non-negative PU learning which indicates that
% \begin{align}
% (1-\pi)E_{p(x|y=-1)}[\mathcal{L}[g(x),-1]]=E_{p(x)}[\mathcal{L}[g(x),-1]]-\pi E_{p(x|y=+1)}[\mathcal{L}[g(x),-1]]\ge0
% \end{align}
% should always holds but sometimes not holds in $\hat{R}_{upu}(g)$. So the non-negative PU Learning loss can be derived as 
% \begin{align}
% \hat{R}_{nnpu}(g)=\frac{\pi}{n_p}\sum_{x_i\in P}\mathcal{L}[g(x_i),+1]+max(0, \frac{1}{n_u}\sum_{x_i\in U}\mathcal{L}[g(x_i),-1]-\frac{\pi}{n_p}\sum_{x_i\in P}\mathcal{L}[g(x_i),-1]).
% \end{align}
\citep{su2021positive} argued that previous approaches struggle with balanced metrics. The objective of unbiased PU learning is to train a classifier that is unbiased when the class distribution of the test data matches that of the unlabeled data, rather than creating a balanced classifier. This can lead to poor performance for one of the classes, especially when $\pi$ is close to 0 or 1. In such cases, even if the model classifies all samples into a single class, achieving high accuracy, it does not meet the goals for real-world applications. Our objective is to learn a balanced classifier, despite the imbalance between positive and negative data in the unlabeled set. Building on the balanced PN learning objective, we aim to address this challenge:
% \begin{align}
% R_{balancedpn}(g)&=\frac{1}{2}E_{p(x|y=+1)}[\mathcal{L}[g(x),+1]]\nonumber\\&+\frac{1}{2}E_{p(x|y=-1)}[\mathcal{L}[g(x),-1]],
% \end{align}
\begin{align}
R_{bpn}(g)=\frac{1}{2}E_{p(x|y=+1)}[\mathcal{L}[g(x),+1]]+\frac{1}{2}E_{p(x|y=-1)}[\mathcal{L}[g(x),-1]],
\end{align}
Through empirical estimation, we can get the balanced PU learning objective:
% \begin{align}
% \hat{R}_{balancedpu}(g)&=\frac{1}{2n_p}\sum_{x_i\in P}\mathcal{L}[g(x_i),+1]\nonumber\\&+\frac{1}{2n_u(1-\pi)}\sum_{x_i\in U}\mathcal{L}[g(x_i),-1]\nonumber\\&-\frac{\pi}{2n_p(1-\pi)}\sum_{x_i\in P}\mathcal{L}[g(x_i),-1].
% \end{align}
\begin{align}
\hat{R}_{bpu}(\hat{g})&=\frac{1}{2n_p}\sum_{x_i\in P}\mathcal{L}[\hat{g}(x_i),+1]+\frac{1}{2n_u(1-\pi)}\sum_{x_i\in U}\mathcal{L}[\hat{g}(x_i),-1]\nonumber\\&-\frac{\pi}{2n_p(1-\pi)}\sum_{x_i\in P}\mathcal{L}[\hat{g}(x_i),-1].
\end{align}
Theoretically, the loss function most related to accuracy is the 0-1 loss. A surrogate loss function that is unbiased with respect to the 0-1 loss should satisfy the condition $\mathcal{L}[t,+1]+\mathcal{L}[t,-1]=1$. We directly set $\mathcal{L}[\hat{g}(x_i),-1]=1-\mathcal{L}[\hat{g}(x_i),+1]$ in $\hat{R}_{bpu}(g)$, yielding a simplified expression:
% \begin{align}
% \hat{R}_{balancedpu}(g)&=\frac{1}{2n_p(1-\pi)}\sum_{x_i\in P}\mathcal{L}[g(x_i),+1]\nonumber\\&+\frac{1}{2n_u(1-\pi)}\sum_{x_i\in U}\mathcal{L}[g(x_i),-1]-\frac{\pi}{2(1-\pi)}
% \end{align}
\begin{align}
\label{bpu}
\hat{R}_{bpu}(g)=\frac{1}{2n_p(1-\pi)}\sum_{x_i\in P}\mathcal{L}[\hat{g}(x_i),+1]+\frac{1}{2n_u(1-\pi)}\sum_{x_i\in U}\mathcal{L}[\hat{g}(x_i),-1]-\frac{\pi}{2(1-\pi)}.
\end{align}
This simple formula illustrates a straightforward yet counterintuitive principle. When the loss function satisfies $\mathcal{L}[t,+1] + \mathcal{L}[t,-1]$ as a constant, directly treating unlabeled samples as negative and training the model using the expected loss for each class ‌supervisely results in a balanced learner. In other words, the model becomes unbiased with respect to the average accuracy metric. Building on this, we further derive the micro-level learning objective for PU learning.
\section{Balanced Fine-Grained PU Learning}
\subsection{Micro-to-Macro Optimization}
In the problems we encounter, the key information that determines a sample as anomalous represents only a small portion of the anomalous samples. Specifically, all local components within the normal samples do not contain anomalous information, while in the anomalous samples, apart from a small amount of local anomalous information, the rest is normal information. We are given a macro dataset $D_{macro}=\{(X_1,Y_1),\dots,(X_{|D_{macro}|},Y_{|D_{macro}|})\}\}$. It can be represented separately as $P_{macro}$ and $N_{macro}$. For a macro sample of length $l$: $X_i=[x_{i1},x_{i2},\cdots,x_{il}]$, $x_{ij}\in\mathbb{R}^d$, $j\in [1,l]$ is the input, and $Y_i=F([y_{i1},y_{i2},\cdots,y_{il}]), y_{ij}\in\{-1, +1\}$ is the output where $F$ is the function that transforms micro labels into macro labels which constitutes a MIL problem~\citep{carbonneau2018multiple}:
\begin{align}
\begin{split}
F([y_{i1},\cdots,y_{il}]) = \left \{
\begin{array}{ll}
    +1,     & \forall j \in\{1,\cdots,l\}, y_{ij}=+1\\
    -1,     &  \exists j \in\{1,\cdots,l\},  y_{ij}=-1
\end{array}
\right.
\end{split}
\end{align}
% 考虑我们的设置，我们需要完成微观的PU Learning，针对我们的任务目标，我们不再追求PU学习模型在细粒度上的性能，我们最终的目标是使在细粒度上训练的模型能尽可能有利于粗粒度上的分类。我们定义模型在粗粒度上的平衡PN学习目标为
We denote $G$ as the macro decision function which is transformed from the micro decision function $g$. Based on the relationship between the micro classifier $f$ and the macro classifier $F$, that is, within a set of macro data, if all micro-components belong to the normal class, the macro label is normal; if there exists at least one anomalous component, the macro label is anomalous. This can be reformulated as another description: if the micro component most inclined towards the anomalous class is anomalous, then the macro label is anomalous; otherwise, the macro label is normal. If such a micro component can be found through $g$ which is the idealized decision function, $G$ can be defined:
% \begin{align}
% \label{G}
% &G([g(x_1),g(x_2),\cdots,g(x_l)])
% \nonumber\\=&\sum_{i=1}^l\mathbb{I}[i==\arg\min_j g_{-1}^*(x_j)]g(x_i)
% \end{align}
\begin{align}
\label{G}
G([g(x_{i1}),g(x_{i2}),\cdots,g(x_{il})])
=\sum_{j=1}^lp(j=\arg\max_k g_{-1}(x_{ik}))g(x_{ij})
% \sum_{j=1}^l\mathbb{I}[i==\arg\max_j g_{-1}^*(x_j)]g(x_i)
\end{align}
Referring to \cref{G}, we aim to find the idealized $p$ which makes $p(j=\arg\max_k g_{-1}(x_{ik}))=\mathbb{I}[j=\arg\max_k g_{-1}(x_{ik})]$. However, since we cannot obtain $g$, during the training process of neural networks, $\hat{g}$ from the neural network often deviates somewhat from the ideal decision function $g$. Consequently, we cannot directly obtain the value of the indicator function $\mathbb{I}[j=\arg\min_k g_{-1}(x_{ik})]$, but we can estimate the probability of the condition being satisfied. So, we use the probability function $\hat{p}$ instead of the function $p$. 
\begin{align}
\label{hatp}
    \hat{p}(x_{ij})=\frac{\exp(\hat{g}_{-1}(x_{ij}))}{\sum_{k=1}^{l} \exp(\hat{g}_{-1}(x_{ik}))}
\end{align}

Given our task, we are no longer pursuing the performance of the PU learning model at the micro level. Our ultimate goal is to ensure that the model trained at the micro level is beneficial for macro learning. We define the unbiased and balanced coarse-grained PN learning objective for the model as
% \begin{align}
%     &\hat{R}_{bfgpn}(g)\nonumber\\=&\frac{1}{2|P_{macro}|}\sum_{X_i \in P_{macro}}\mathcal{L}[G([g(x_{i1}),\cdots,g(x_{il})]),+1]
% \nonumber\\+&\frac{1}{2|N_{macro}|}\sum_{X_i \in N_{macro}}\mathcal{L}[G([g(x_{i1}),\cdots,g(x_{il})]),-1]
% \end{align}
% \begin{align}
%     &\hat{R}_{bfgpn}(g)=\frac{1}{2|P_{macro}|}\sum_{X_i \in P_{macro}}\mathcal{L}[G([g(x_{i1}),\cdots,g(x_{il})]),+1]
% \nonumber\\+&\frac{1}{2|N_{macro}|}\sum_{X_i \in N_{macro}}\mathcal{L}[G([g(x_{i1}),\cdots,g(x_{il})]),-1]
% \end{align}
\begin{align}
\label{cgpn}
   \hat{R}_{cgpn}(\hat{g})=&\frac{1}{2\cdot|P_{macro}|\cdot l}\sum_{X_i \in P_{macro}}\mathcal{L}[G([\hat{g}(x_{i1}),\cdots,\hat{g}(x_{il})]),+1]
\nonumber\\+&\frac{1}{2\cdot|N_{macro}|\cdot l}\sum_{X_i \in N_{macro}}\mathcal{L}[G([\hat{g}(x_{i1}),\cdots,\hat{g}(x_{il})]),-1]
\end{align}
% 其中G函数用于实现细粒度决策函数向粗粒度决策函数的转换。根据细粒度的分类器f和粗粒度分类器F的关系，即一项粗粒度数据中，如果所有细粒度成分都属于正类，则粗粒度为正，如果存在一个负类，则粗粒度为负，这可以转换为另一种描述：如果细粒度上最偏向负类的成分为负，则粗粒度为负，否则粗粒度为正，即定义
% 上述G函数的一个隐含假设是g(x)是足够精确的，可以准确反映不同细粒度成分偏向正类或负类的程度，然而在神经网络的训练过程中，g(x)通常离理想的决策函数有一定的差异，对于最接近负类的样本的判断可能出现错误，这是很容易引发风险的。而指示函数可视为取值为0或1的加权函数，在g(x)不够精确时，我们使用更平滑的加权函数可以取得更好的效果，即使用权重函数
% The implicit assumption of the function $G$ mentioned above is that $g(x)$ is accurate enough to reflect the degree to which different micro samples lean towards the positive or negative class. However, during the training process of neural networks, $g(x)$ often deviates somewhat from the ideal decision function. 
Based on \cref{bpu,G,hatp,cgpn}, through the transformation of the learning paradigm, we derive a loss function that enables unbiased and balanced optimization of macro-level performance directly at the micro level which is called balanced fine-grained PU learning loss:
% \begin{align}
%     R_{bfgpu}(g)=&\frac{1}{2|P_{macro}|}\sum_{X_i\in P_{macro}}\sum_{j=1}^{l}p_{\mathbb{I}}(x_{ij})\mathcal{L}[g(x_{ij}),+1]
% \nonumber\\+&\frac{1}{2|N_{macro}|}\sum_{X_i\in N_{macro}}\sum_{j=1}^{l}p_{\mathbb{I}}(x_{ij})\mathcal{L}[g(x_{ij}),-1]
% \end{align}
\begin{align}
\label{bfgpu}
    \hat{R}_{cgpn}(\hat{g})=\hat{R}_{bfgpu}(\hat{g})=&\frac{1}{2|P_{micro}|}\sum_{x_{ij}\in P_{micro}}\hat{p}(x_{ij})\mathcal{L}[\hat{g}(x_{ij}),+1]
\nonumber\\+&\frac{1}{2|U_{micro}|}\sum_{x_{ij}\in U_{micro}}\hat{p}(x_{ij})\mathcal{L}[\hat{g}(x_{ij}),-1]
\end{align}
\subsection{Pseudo Labels Based on Macro Information}
% 在经过PU Learning训练后，模型对于正负样本已经具备了一定的区分力，可以考虑进一步从无标注数据中获取伪标注进行进一步的PN Learning训练。且在前一阶段我们只用到了样本的微观信息，而忽略了样本的宏观信息，即微观样本对宏观样本的归属关系。而在新的阶段，我们可以考虑利用好这些宏观信息，对于一条宏观负样本，由于我们知道其中至少包含一个微观负样本，所以可以直接将其中最偏向负类的微观样本标记为负样本，另外为了保障学习器的平衡性，我们同时将其中最偏向正类的微观样本标记为正样本，这样我们就得到了相同数量的带有伪标注的正负样本，用于进一步训练模型即可保障模型性能改善的同时对于正负样本是平衡的。
After an epoch of PU learning, the model has acquired some discriminative ability for normal and anomalous samples. Further training with PN learning can be considered by obtaining pseudo-labels from unlabeled data. In the previous stage, we only utilized the micro information of the samples, neglecting the macro information. In the new stage, we can leverage this macro information. For an anomalous macro sample, since we know it contains at least one anomalous micro sample, we can directly label the most inclined-to-anomalous micro sample as an anomalous sample. Additionally, to ensure the balance of the learner, we simultaneously label the most inclined-to-normal micro sample as a normal sample. This way, we obtain an equal number of positive and negative pseudo-labeled samples for further model training. Expressed symbolically as:
\begin{alignat}{2}
\label{18}
N_{pse}=\{(x_{ij},-1), j=\arg\max_k \hat{g}_{-1}(x_{ik}) ,X_i\in N_{macro}\}\\
\label{19}
P_{pse}=\{(x_{ij},+1), j=\arg\min_k \hat{g}_{-1}(x_{ik}) ,X_i\in N_{macro}\} 
\end{alignat}
Then we use $N_{pse}$ and $P_{pse}$ to further train the model. This ensures the improvement while maintaining the balance between positive and negative samples. The loss function is:
% \begin{spreadlines}{-1em}
\begin{align}
\label{20}
    \hat{R}_{pse}(\hat{g})=\sum_{(x_{i},y_{i})\in N_{pse}+P_{pse}}\mathcal{L}[\hat{g}(x_{i}),y_i]
    % \vskip-0.4in
\end{align}
% \end{spreadlines}
% 在所有训练过程都结束后，我们会将模型投入测试，用于对异常数据的检测，而在检测过程中需要设置恰当的阈值，通过决策函数和阈值的比较确定最终的标注。通常我们会默认阈值为0.5，但是我们发现完全可以利用类别分布$\pi$确认更精确的阈值。
\subsection{Adjusted Decision Threshold (ADT)}
After the training is completed, we deploy the model for testing. During the learning process, it is necessary to set an appropriate threshold. The final label is determined by comparing the decision function with the threshold. Typically, a default threshold of $0.5$ is assumed, but a more precise threshold can be determined by using the normal label distribution $\pi=1-\frac{1}{(\sigma_{micro}+1)\cdot (\sigma_{macro}+1)}$. Here, $\sigma_{micro}=(l-1)$ and $\sigma_{macro}=\frac{|P_{macro}|}{|N_{macro}|}$ denote the imbalance ratios of the MIL problem at the micro and macro levels, respectively. Since we have access to the decision function values $\hat{g}(x)$, for all micro-unlabeled samples, it is sufficient to sort $\hat{g}(x)_{-1}$ in ascending order. After sorting, we can choose the position that corresponds to $\pi$ as the threshold $T$:
\begin{align}
\label{21}
    T=sort([\hat{g}_{-1}(x_i), x_i \in U_{micro}])[\lfloor |U_{micro}|\cdot\pi\rfloor]
\end{align}
This approach helps determine an accurate threshold because the decision function $g(x)$ reflects the conditional probability $p(y|x)$. $p(y|x)$ remains consistent between training and testing data. Therefore, the threshold obtained on the unlabeled data is highly applicable to the test data. The overall process of the algorithm can be seen in \cref{algorithm}.

\section{Experiments}
% We conducted numerous experiments to validate the effectiveness. The experimental datasets comprised 3 datasets with small imbalanced ratios, 2 datasets with large imbalanced ratios, and 1 dataset from a real-world scenario. Furthermore, we conducted experiments under extreme imbalance conditions, both at the macro and micro levels. Finally, to validate the importance of each module and the stability under different hyperparameters, we conducted ablation experiments and sensitive analysis.
% \begin{figure}[htbp]
% % \vskip 0.2in
% \begin{center}
% \centerline{\includegraphics[scale=0.5]{Fig/sst25.png}}
% \caption{Experiments on SST-2\\ when $\sigma_{macro}=5$}
% \label{sst25}
% \end{center}
% % \vskip -0.4in
% \end{figure}
    % \end{minipage}%
    % \begin{minipage}{0.5\textwidth} % 第二列，占页面宽度的50%
    % \end{minipage}
\subsection{Experimental Settings}
\label{settings}
% 对于文本模态的数据我们主要使用Roberta作为基础模型（针对基础模型的消融实验除外），对于图像模态的数据我们使用ResNet18作为基础模型队所有算法进行公平对比。我们统一使用Adam作为优化器
For textual data, we primarily employ RoBERTa~\citep{liu2019roberta} as the backbone model (except for ablation studies concerning the backbone itself). For image data, we utilize ResNet-18 as the backbone to ensure a fair comparison across all algorithms. We uniformly adopt Adam as the optimizer ~\citep{kingma2014adam}, with a learning rate set to $10^{-5}$, and CosineAnnealingLR as the learning rate scheduler. The number of epochs was set to $5$, batch size to $16$. We denote $\sigma_{micro}$ as the imbalance ratio at the micro level, which represents the positive-negative ratio of micro samples in negative macro samples. $\sigma_{macro}$ represents the imbalance ratio at the macro level, which is the ratio of positive macro samples to negative macro samples. We set $\lambda_{bfgpu}=1/\pi$ and $\lambda_{pse}=1$ in all experiments. For micro-level PU learning, $\pi=\frac{\sigma_{micro}}{\sigma_{micro}+1}$. The algorithms were implemented using the PyTorch framework ~\citep{paszke2019pytorch}. We used average accuracy and F1 score, two commonly used evaluation metrics in imbalanced learning. Regarding the F1 score, when dealing with class imbalance problems, we typically focus on the performance of the minority (anomalous) class. We also compared the algorithm performance as $\sigma_{micro}$ varies, using the area under the curve $AUC_{AvgAcc}$ and $AUC_{F1}$. 4 A800 GPUs are used for all experiments.
% and all experiments were repeated $3$ times with random seeds $[0, 1, 2]$ to obtain the mean and variance of performance
% For public datasets, a macro-level normal sample is composed of $\sigma_{micro}+1$ positive instances, while a macro-level anomalous sample is composed of $\sigma_{micro}$ positive instances and $1$ negative instanceIn the experiments in \cref{public,real,ablation}, $\sigma_{macro}$ was set to $1$, while in the experiments in \cref{both}, $\sigma_{macro}$ was set to $5$. $\sigma_{micro}$ was set to $[2, 4, 6, 8, 10]$ for short texts and $[2, 3, 4, 5]$ for moderate texts. In \cref{real}, $\sigma_{micro}$ varied inconsistently averaging around $25$.
% 15 algorithms were used for comparison, including 1 macro classification algorithm, 1 macro anomaly detection algorithm~\citep{ruff2019deep}, 2 micro anomaly detection algorithms~\citep{ruff2018deep,ruff2019deep}, 3 multi-instance learning algorithms~\citep{angelidis2018multiple,ilse2018attention,perini2023learning}, 4 positive-unlabeled (PU) learning algorithms~\citep{du2015convex,kiryo2017positive,su2021positive,zhu2023robust}, and their enhanced versions after applying ADT we proposed.
\subsection{Compared Methods}
\label{Comp}
Our comparative algorithms consist of:
\textbf{Macro Supervised}: Conventional macro binary classification.
\textbf{Macro DeepSAD}: Supervised macro AD using DeepSAD~\citep{ruff2019deep}.
\textbf{Macro Imbalanced Learning}: Under sampling and Over sampling.
\textbf{MIL}: MIL algorithms used to address inexact supervision, using five multi-instance learning algorithms MIL-MaxPooling, MIL-TopkPooling (k was set to 3), MIL-Attention ~\citep{ilse2018attention}, MIL-FGSA ~\citep{angelidis2018multiple}, and MIL-PReNET~\citep{pang2023deep}.
\textbf{Micro DeepSAD}: Supervised micro AD, using DeepSAD ~\citep{ruff2019deep} and treating unlabeled samples as negative class. 
\textbf{Micro DeepSVDD}: Unsupervised micro AD, using DeepSVDD ~\citep{ruff2018deep} which is an One-Class Classification (OCC) method using only positive samples. 
\textbf{Micro PU Learning}: Utilizing four types of loss functions uPU ~\citep{du2015convex}, nnPU ~\citep{kiryo2017positive}, balancedPU ~\citep{su2021positive}, and robustPU~\citep{zhu2023robust} for micro-level PU learning. \textbf{LLMs}: GPT-4-turbo~\citep{achiam2023gpt}, Qwen3~\citep{yang2025qwen3}, Deepseek v3~\citep{liu2024deepseek}, Gemini2.5~\citep{comanici2025gemini}.
% \textbf{Enhanced Versions of Micro PU Learning}: Incorporating our proposed ADT to Micro PU learning, denoted as Micro uPU+ADT, Micro nnPU+ADT, Micro balancedPU+ADT, and Micro robustPU+ADT.
% \begin{table}[htbp]
% \centering
% \captionof{table}{Experimental Results on the Customer Service Quality Inspection Dataset in the Real Application.}
% \label{real_table}
% \begin{tabular}{c c c c}
% \hline\hline
% &Method &AvgAcc&F1 Score\\
%     \hline\hline
%     \multirow{2}{*}{Macro}&PN&$52.38\pm4.12$&$8.33\pm14.33$\\
%     % Coarse-Grained AD&$75.95\pm1.52$&$73.60\pm0.26$&$70.58\pm0.48$&$68.61\pm0.22$&$61.41\pm6.18$\\
%     &DeepSAD&$47.61\pm4.12$&$6.67\pm11.55$\\\hline
%     \multirow{9}{*}{Macro}&DeepSAD&$58.33\pm5.46$&$63.12\pm4.89$\\
%     &DeepSVDD&$50.00\pm0.00$&$0.00\pm0.00$\\
%     &MIL-Attention&$54.76\pm11.48$&$53.27\pm8.72$\\
%     &MIL-FGSA&$53.57\pm7.14$&$54.14\pm9.12$\\
%     &MIL-PReNET&$59.52\pm2.06$&$59.05\pm0.86$\\
%     &uPU&$50.00\pm0.00$&$0.00\pm0.00$\\
%     &nnPU&$50.00\pm0.00$&$0.00\pm0.00$\\
%     &balancedPU&$50.00\pm0.00$&$0.00\pm0.00$\\
%     &robustPU&$50.00\pm0.00$&$0.00\pm0.00$\\
%     % &uPU+ADT&$48.81\pm8.25$&$46.67\pm15.77$\\
%     % &nnPU+ADT&$48.81\pm8.25$&$46.67\pm15.77$\\
%     % &balancedPU+ADT&$52.38\pm10.91$&$45.45\pm18.43$\\
%     % % &+ADT&&\\ \multirow{2}{*}{ \multirow{2}{*}{
%     % &robustPU+ADT&$52.38\pm5.45$&$51.33\pm6.28$\\
%     % &+ADT&&\\\multirow{2}{*}{\multirow{2}{*}{
% &BFGPU&\bm{$71.43\pm9.45$}&\bm{$75.31\pm5.84$}\\
%     \hline\hline
%     \end{tabular}
%     \end{table}
\begin{figure}[h!]
\vskip-0.1in
    \centering
    \begin{minipage}{0.48\textwidth} % 第一列，占页面宽度的50%
    % \centering
    % \includegraphics[scale=0.3]{Fig/macro5.png} % 请替换为你的图片文件名
    % \label{sst25}
    % \caption{The figure presents the AvgAcc of different algorithms under both micro and macro imbalance on SST-2.}
    \centering
\captionof{table}{The comparison on the IDC dataset.}
\label{idc_table}
\begin{tabular}{c c c c}\hline\hline
&Method&	AvgAcc&	F1 Score\\\hline\hline
\multirow{2}{*}{Macro}&Supervised&	58.70&	57.13\\
&DeepSAD&	44.30&	43.20\\\hline
\multirow{5}{*}{MIL}&Maxpooling&	65.55&	62.39\\
&Topkpooling&	63.74&	63.82\\
&Attention&	56.03&	49.10\\
&FGSA&	60.22&	59.86\\
&PReNET&	49.30&	38.46\\\hline
\multirow{7}{*}{Micro}&DeepSAD&	61.94&	58.81\\
&DeepSVDD&	52.15&	49.10\\
&uPU&	56.88&	56.36\\
&nnPU&	62.66&	61.00\\
&balancePU&	57.33&	55.73\\
&robustPU&	64.38&	63.16\\
&BFGPU&	\textbf{73.87}&	\textbf{73.21}\\
\hline\hline
\end{tabular}
    % \caption{Experiments when $\sigma_{macro}=5$}
    \end{minipage}%
    \hfill
    \begin{minipage}{0.48\textwidth} % 第二列，占页面宽度的50%
    \centering
    \captionof{table}{The comparison on the CSQI dataset.}
    \label{real_table}
% \begin{tabular}{c c c c}
% \hline\hline
% &Method &AvgAcc&F1 Score\\
%     \hline\hline
%     \multirow{2}{*}{Macro}&PN&$52.38\pm4.12$&$8.33\pm14.33$\\
%     % Coarse-Grained AD&$75.95\pm1.52$&$73.60\pm0.26$&$70.58\pm0.48$&$68.61\pm0.22$&$61.41\pm6.18$\\
%     &DeepSAD&$47.61\pm4.12$&$6.67\pm11.55$\\\hline
%     \multirow{3}{*}{MIL}
%     &Attention&$54.76\pm11.48$&$53.27\pm8.72$\\
%     &FGSA&$53.57\pm7.14$&$54.14\pm9.12$\\
%     &PReNET&$59.52\pm2.06$&$59.05\pm0.86$\\\hline
%     \multirow{7}{*}{Micro}
%     &DeepSAD&$58.33\pm5.46$&$63.12\pm4.89$\\
%     &DeepSVDD&$50.00\pm0.00$&$0.00\pm0.00$\\
%     &uPU&$50.00\pm0.00$&$0.00\pm0.00$\\
%     &nnPU&$50.00\pm0.00$&$0.00\pm0.00$\\
%     &balancedPU&$50.00\pm0.00$&$0.00\pm0.00$\\
%     &robustPU&$50.00\pm0.00$&$0.00\pm0.00$\\
%     % &uPU+ADT&$48.81\pm8.25$&$46.67\pm15.77$\\
%     % &nnPU+ADT&$48.81\pm8.25$&$46.67\pm15.77$\\
%     % &balancedPU+ADT&$52.38\pm10.91$&$45.45\pm18.43$\\
%     % % &+ADT&&\\ \multirow{2}{*}{ \multirow{2}{*}{
%     % &robustPU+ADT&$52.38\pm5.45$&$51.33\pm6.28$\\
%     % &+ADT&&\\\multirow{2}{*}{\multirow{2}{*}{
% &BFGPU&\bm{$71.43\pm9.45$}&\bm{$75.31\pm5.84$}\\
%     \hline\hline
%     \end{tabular}
\begin{tabular}{c c c c}
\hline\hline
&Method &AvgAcc&F1 Score\\
    \hline\hline
    \multirow{2}{*}{Macro}&Supervised&$52.38$&$8.33$\\
    % Coarse-Grained AD&$75.95\pm1.52$&$73.60\pm0.26$&$70.58\pm0.48$&$68.61\pm0.22$&$61.41\pm6.18$\\
    &DeepSAD&$47.61$&$6.67$\\\hline
    \multirow{5}{*}{MIL}
    &Maxpooling&$48.81$&$46.67$\\
    &Topkpooling&$52.38$&$45.45$\\
    &Attention&$54.76$&$53.27$\\
    &FGSA&$53.57$&$54.14$\\
    &PReNET&$59.52$&$59.05$\\\hline
    \multirow{7}{*}{Micro}
    &DeepSAD&$58.33$&$63.12$\\
    &DeepSVDD&$50.00$&$0.00$\\
    &uPU&$50.00$&$0.00$\\
    &nnPU&$50.00$&$0.00$\\
    &balancedPU&$50.00$&$0.00$\\
    &robustPU&$50.00$&$0.00$\\
    % &uPU+ADT&$48.81\pm8.25$&$46.67\pm15.77$\\
    % &nnPU+ADT&$48.81\pm8.25$&$46.67\pm15.77$\\
    % &balancedPU+ADT&$52.38\pm10.91$&$45.45\pm18.43$\\
    % % &+ADT&&\\ \multirow{2}{*}{ \multirow{2}{*}{
    % &robustPU+ADT&$52.38\pm5.45$&$51.33\pm6.28$\\
    % &+ADT&&\\\multirow{2}{*}{\multirow{2}{*}{
&BFGPU&\bm{$71.43$}&\bm{$75.31$}\\
    \hline\hline
    \end{tabular}
    \end{minipage}
    \vskip-0.2in
\end{figure}
% \begin{figure}[htbp]
%     \centering
%     % \begin{minipage}{0.5\textwidth} % 第一列，占页面宽度的50%
%     \centering
%     \includegraphics[scale=0.43]{Fig/macro5.png} % 请替换为你的图片文件名
%     \caption{Experimental Results on the SST-2 Dataset with Imbalance at Both Macro and
% Micro Levels}
%     \label{sst25}
%     \end{figure}
% \begin{table}[ht]
% \centering
% \small
% \caption{Experiments on Customer Service Quality Inspection Dataset in the Real Application.}
% \label{real_table}
% % \vskip 0.15in
% \vskip -0.25in
% \end{table}
% \subsection{Compared Methods}
\subsection{Experiments with the IDC Dataset}
\label{IDC}
We evaluated various methods with a standard MIL dataset IDC, which is a medical image classification dataset primarily used for breast cancer detection research \citep{bolhasani2020histopathological}. This dataset contains patches of breast tissue images labeled according to the presence of invasive ductal carcinoma. In the dataset setup, each bag is a patient and the $\sigma_{macro}$ is around $3$. Each instance is a 50×50 pixel image patch, and each bag contains 4 instances. The results are shown in \cref{idc_table}.
\subsection{Experiments with the CSQI Dataset}
\label{real}
We evaluated various methods with a real-world CSQI dataset depicted in \cref{Data}. The dataset consists of instances where service personnel were flagged for substandard performance during service. In CSQI, anomalous samples are extremely scarce, with only 24 dialogue sessions, while normal samples are abundant. we set $\sigma_{macro}=100$ and randomly sampled 2,400 normal sessions for the experiments. The $\sigma_{micro}$ varied inconsistently, averaging around $25$. The results are shown in \cref{real_table}. 
\subsection{Experiments with Synthetic Datasets}
\label{public}
% 我们分别在imdb、amazon、yelp三个文本长度适中的数据集上进行了$\sigma_{micro}$取值为[2,3,4,5]的实验，实验结果如表一所示
Since real datasets cannot control the imbalance ratios $\sigma_{macro}$ and $\sigma_{micro}$, we use sentiment analysis datasets to synthesize datasets with different imbalance ratios to explore the capability of various algorithms in handling dual imbalance problems. For example, we randomly combine 5 positive samples and 1 negative sample from the original dataset to synthesize a positive sample with $\sigma_{micro}=5$. For short-text ones SST-2 ~\citep{socher2013recursive} and Sentiment140 ~\citep{go2009twitter}, $\sigma_{micro}$ was set to $[2, 4, 6, 8, 10]$. For long-text datasets IMDB~\citep{maas2011learning} and Amazon, $\sigma_{micro}$ was set to $[2, 3, 4, 5]$, the remaining experimental results can be found in. All experiments were repeated $3$ times using random seeds $[0, 1, 2]$. The results can be found in \cref{sst2_table,sst2_f1_table,imdb_f1_table,sentiment_table,amaozn_table}.
\begin{table*}[htbp]
\centering
\small
\caption{The table presents the AvgAcc of various algorithms under varying values of $\sigma_{\text{micro}} \in \{2, 4, 6, 8, 10\}$, along with the estimated AUC concerning changes in AvgAcc on the SST-2 Dataset.}
\label{sst2_table}
% \vskip 0.15in
\resizebox{1\textwidth}{!}{\begin{tabular}{c c c c c c c c}
\hline\hline
&\multirow{2}{*}{Methods} & \multicolumn{5}{c}{AvgAcc} & \multirow{2}{*}{$AUC_{AvgAcc}$}\\\cline{3-7}
&&2&4&6&8&10&\\
    \hline\hline
    \multirow{2}{*}{Macro} &Supervised&$83.15\pm0.98$&$78.38\pm1.02$&$76.57\pm3.42$&$71.38\pm9.29$&$73.26\pm2.51$&$76.55$\\
    % Coarse-Grained Sad&$79.47\pm0.34$&$73.27\pm4.66$&$72.46\pm3.29$&$56.92\pm3.11$&$53.88\pm1.45$\\
    &DeepSAD&$76.52\pm0.23$&$59.41\pm6.75$&$52.17\pm1.02$&$49.69\pm3.47$&$50.39\pm2.19$&$57.64$\\\hline
    \multirow{5}{*}{MIL} 
    &MaxPooling&$85.82\pm1.78$&$66.83\pm6.44$&$61.84\pm10.31$&$52.83\pm4.90$&$56.20\pm10.74$&$64.70$\\
    &TopkPooling&$-$&$75.25\pm5.35$&$75.85\pm2.74$&$68.87\pm3.27$&$64.73\pm12.81$&$71.18$\\
    &Attention&$80.76\pm1.57$&$70.96\pm3.72$&$57.49\pm12.35$&$50.00\pm0.00$&$51.16\pm2.01$&$60.07$\\
    &FGSA&$68.05\pm16.87$&$50.00\pm0.00$&$50.00\pm0.00$&$50.00\pm0.00$&$50.00\pm0.00$&$53.61$\\
    &PReNET	&$78.82\pm6.43$&$67.33\pm3.25$&$66.49\pm3.32$&$56.92\pm6.07$&$60.47\pm3.49$&$66.00$\\\hline
\multirow{7}{*}{Micro}     &DeepSAD&$61.33\pm5.75$&$51.16\pm1.53$&$48.79\pm1.49$&$49.06\pm0.77$&$49.22\pm0.55$&$51.91$\\
    &DeepSVDD&$48.62\pm1.93$&$50.00\pm2.97$&$49.52\pm0.42$&$49.06\pm0.94$&$50.00\pm1.16$&$49.44$\\&uPU&$80.02\pm1.92$&$71.29\pm3.86$&$69.81\pm2.80$&$67.92\pm4.00$&$53.88\pm2.74$&$68.58$\\
    &nnPU&$81.49\pm0.00$&$50.17\pm0.00$&$50.00\pm0.00$&$50.00\pm0.00$&$50.00\pm0.00$&$56.33$\\
    &balancedPU&$77.90\pm7.82$&$75.74\pm8.49$&$65.70\pm11.42$&$54.72\pm3.85$&$57.75\pm10.96$&$66.36$\\
    &robustPU&$86.28\pm1.39$&$75.91\pm9.20$&$74.15\pm6.16$&$61.64\pm11.80$&$55.42\pm9.40$&$70.68$\\
    % &uPU+ADT&$82.69\pm2.02$&$76.07\pm1.82$&$71.26\pm3.36$&$64.47\pm0.89$&$61.24\pm4.78$&$71.15$\\
    % &nnPU+ADT&$87.02\pm0.60$&\bm{$83.83\pm1.42$}&$79.23\pm1.49$&$75.16\pm2.48$&$74.03\pm3.95$&$79.85$\\
    % &balancedPU+ADT&$79.83\pm6.66$&$77.23\pm5.73$&$68.84\pm11.79$&$66.04\pm5.82$&$72.09\pm1.64$&$72.81$\\
    % &robustPU+ADT&$86.00\pm1.84$&$74.09\pm20.01$&$78.02\pm2.33$&$72.96\pm8.02$&$61.63\pm14.52$&$74.54$\\
    &BFGPU&\bm{$88.40\pm0.68$}&\bm{$82.51\pm0.62$}&\bm{$82.13\pm0.90$}&\bm{$79.56\pm1.60$}&\bm{$82.56\pm1.64$}&\bm{$83.03$}\\
    \hline\hline
\end{tabular}}
\vskip -0.15in
\end{table*}
% We present the AvgAcc results on the SST-2 and IMDB datasets in \cref{sst2_table,imdb_table}. The remaining experimental results can be found in \cref{sst2_f1_table,imdb_f1_table,sentiment_table,amaozn_table}, which are in the appendix.
% 由于空间限制，我们仅在正文中展示了SST-2数据集和Imdb数据集在平均准确率这一指标上的结果，剩余实验结果可见附录。
% The experimental results are shown in \cref{amaozn_table,imdb_table,yelp_table} in the appendix.
% \subsection{Experiments with Large Imbalance Ratio}
% \label{large}
% We conducted experiments with $\sigma_{micro}$ values of 
% The results are shown in \cref{sst2_table,imdb_table} and \cref{sentiment_table,amaozn_table} in the appendix.
% \subsection{Experiments with Imbalance at Both Level}
% \label{both}
We considered a more extreme scenario where the normal-anomalous ratio at not only the micro but also the macro level is imbalanced. This lead to the anomalous information accounting for only $\frac{1}{(1+\sigma_{micro})\cdot(1+\sigma_{macro})}$. However, BFGPU still achieved outstanding performance. We set $\sigma_{macro}$ to 5 and 10 and conducted the experiments. We plotted the curve of average accuracy varying using synthetic SST-2 as shown in fig 3 and 4. The experimental results are shown in \cref{sst25_table,sst210_table}. 
\begin{figure}[h!]
    \centering
    \begin{minipage}{0.48\textwidth} % 第一列，占页面宽度的50%
    \centering
    \includegraphics[scale=0.3]{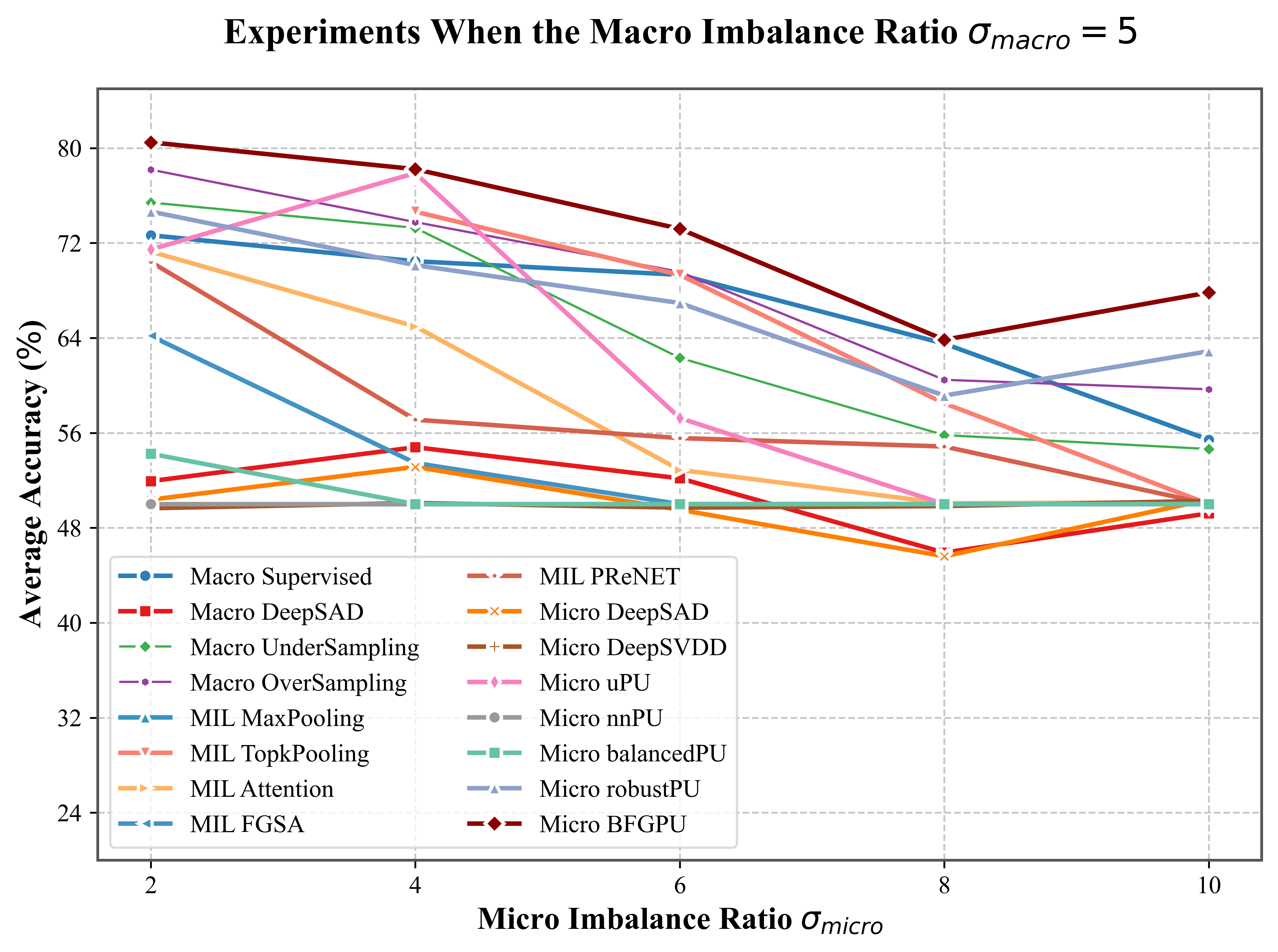} % 请替换为你的图片文件名
    \label{sst25}
    \caption{The figure presents the AvgAcc of various algorithms when $\sigma_{macro}=5$.}
    \end{minipage}
    \hfill
    \begin{minipage}{0.48\textwidth} % 第一列，占页面宽度的50%
    \label{sst210}
    \centering
    \includegraphics[scale=0.3]{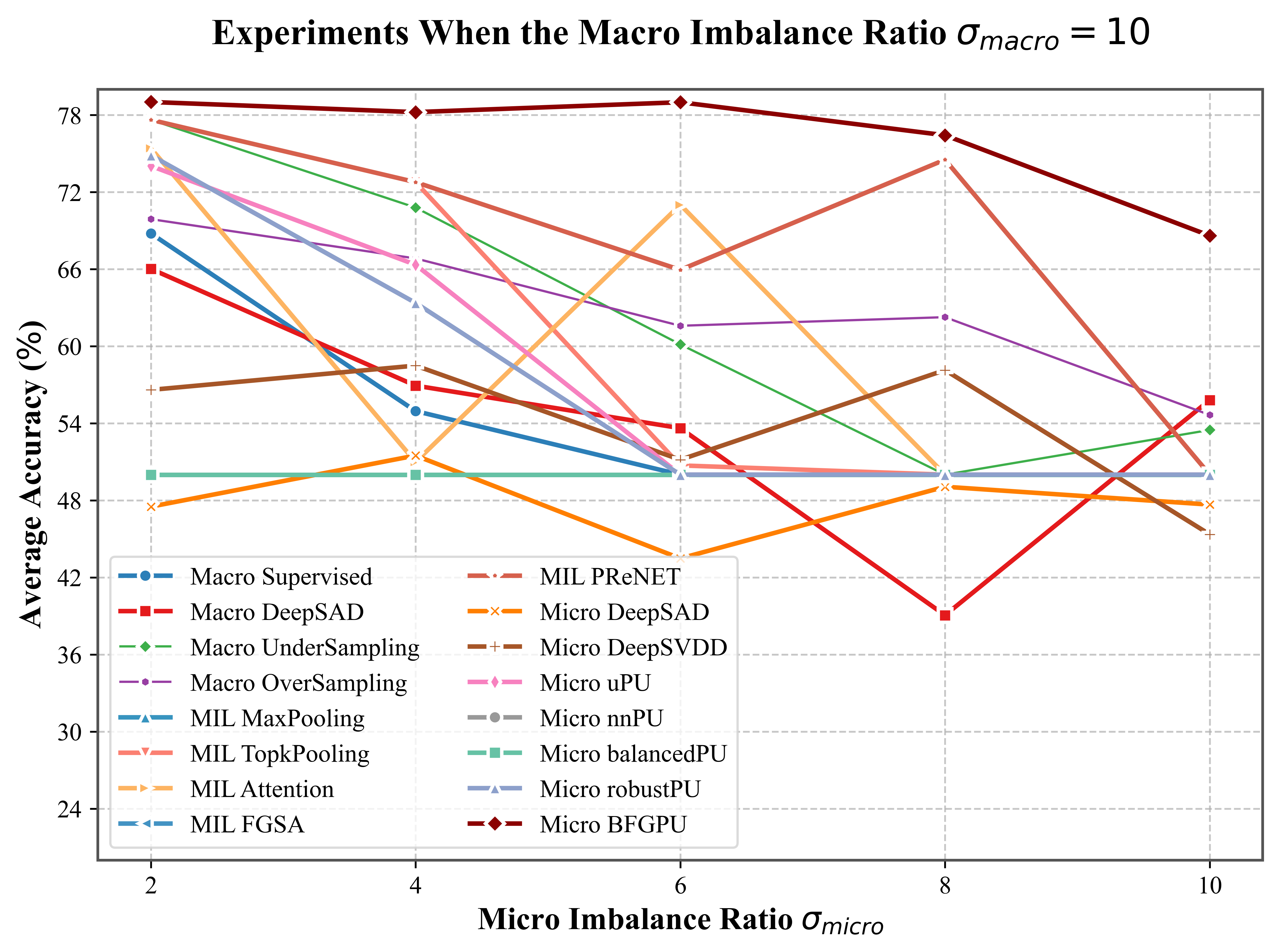} % 请替换为你的图片文件名
    \caption{The figure presents the AvgAcc of various algorithms when $\sigma_{macro}=10$.}
    \end{minipage}
    \vskip -0.2in
\end{figure}
\subsection{Compared with LLMs}
We maximized the $\sigma_{micro}$ in the synthetic datasets, setting it to 5 for IMDB and Amazon, and to 10 for SST-2 and Sentiment140, to evaluate whether the model can accurately identify samples with slight anomalies. This further enabled a comparison of the "needle-in-a-haystack" capability between BFGPU and current mainstream LLMs. We used a consistent prompt: "You are a sentiment classification model; output `positive' if there is no negative sentiment in the paragraph, and `negative' if negative sentiment exists. You must only output `positive' or `negative' without any additional content." The experimental results, shown in \cref{LLM}, confirm that BFGPU is currently more suitable than LLMs themselves as a reward model in the RLHF process.
\begin{table}[htbp]
\vskip-0.1in
\centering
\captionof{table}{The table presents the performance compared with LLMs with maximum $\sigma_{micro}$.}
\label{LLM}
\begin{tabular}{c c c c c c c c c}\hline\hline
\multirow{2}{*}{Model}&\multicolumn{2}{c}{IMDB}&\multicolumn{2}{c}{Amazon}&\multicolumn{2}{c}{SST-2}&\multicolumn{2}{c}{Sentiment140}\\\cline{2-9}
&AvgAcc&	F1 Score&AvgAcc&	F1 Score&AvgAcc&	F1 Score&AvgAcc&	F1 Score\\\hline\hline
GPT-4-turbo&	64.71&71.43&	79.41&86.27&	70.45&	74.95&	55.88&59.46\\
Qwen3&	67.65&66.67&	82.35&	86.36&	67.86&	64.52&	64.71 &	62.50\\
Deepseek v3&	55.88&	48.28&67.65&71.79&	57.58&	56.25&	61.76&	55.17\\
Gemini2.5&	64.71&62.50&79.41	&85.11&	63.09&	74.42&	58.82 &61.11\\
BFGPU&	\bm{$83.64$}&	\bm{$84.61$}&	\bm{$85.57$}&	\bm{$86.50$}&	\bm{$82.56$}&	\bm{$84.41$}&	\bm{$66.45$}&	\bm{$66.40$}\\\hline\hline
\end{tabular}
\vskip-0.3in
\end{table}
% 我们在合成数据集将微观不平衡比例调至最大，即IMDB和Amazon设为5，SST-2和Sentiment140设为10，评估模型是否能否准确找出具有轻微异常的样本，从而进一步对比了BFGPU和目前主流大语言模型的海底捞针的能力，我们使用了统一的prompt：“你是一个情感分类模型，positive表示段落中不存在负面情感，negative表示段落中存在负面情感，你只能输出positive或negative，不输出其他内容”，实验结果见表4，验证了目前BFGPU相较LLMs本身更加适合作为RLHF环节中的奖励模型。
\subsection{Ablation Study and Sensitivity Analysis}
\label{ablation}
% 为了验证我们提出的算法的每一部分的必要性，我们进一步进行了消融实验，除了全面的算法外，我们考虑了另外三种设致：1.不使用加权的细粒度PU学习损失函数；2.不使用伪标注进行进一步训练；3.不使用调整后的阈值。我们在$\sigma_{macro}=5$的设置下在sst-2数据集上进行了实验，$\sigma_{micro}$被设置为$[2,4,6,8,10]$，实验结果如表1所示
We conducted some ablation experiments. a) To validate the necessity of each component of BFGPU, we considered 3 settings: Not using the BFGPU learning loss function; Not using pseudo-labeling for further training; Not using adjusted thresholds. The results are shown in \cref{ablation_table2,ablation_table1}. b) To verify that our results are not limited by the choice of backbone model, we conducted further evaluations using DeBERTa \cref{Deberta}. The results are shown in \cref{Deberta}. c) To further demonstrate the superiority of the BFGPU loss function, we compared its performance against other PU learning methods when integrated with our proposed pseudo-labeling and ADT strategies. The results are shown in \cref{ADT SST2}.

% \subsection{Sensitivity Analysis}
BFGPU mainly has two hyperparameters, $\lambda_{bfgpu}$ and $\lambda_{pse}$. To validate the stability under different hyperparameter settings, we conducted a sensitivity analysis using the SST-2 dataset. The results of $\lambda_{bfgpu}$ and $\lambda_{pse}$ are shown in \cref{sen1} and \cref{sen2} respectively. 
\section{Theoretical Argumentation}
\label{theorem}
We conducted analyses of the generalization error bounds for coarse-grained PN learning, MIL, and balanced fine-grained PU learning, identifying their respective applicability ranges. We also pointed out the necessity of adopting the micro PU learning paradigm, especially when the abnormal content is low. The theoretical analysis below is based on two assumptions ~\citep{niu2016theoretical}:
\begin{assumption}
There is a constant $C_{\mathcal{G}} > 0$ such that: $\mathcal{R}_{n,q}(\mathcal{G})\le C_{\mathcal{G}/\sqrt{n}}$ where $\mathcal{R}_{n,q}(\mathcal{G})$ is the Rademacher complexity of the function space $\mathcal{G}$ for $n$ samples from any marginal distribution $q(x)$.
\end{assumption}
% \begin{align}
% \end{align}
\begin{assumption}
The loss function $L$ satisfies the symmetric condition and $\alpha_L$-Lipschitz continuity:
% \begin{align}
%     l(t,+1)+l(t,-1)&=1,\nonumber\\ 
%     |l(t_1,y)-l(t_2,y)|&\le L_l|t_1-t_2|
% \end{align}
\begin{align}
    L(t,+1)+L(t,-1)=1, |L(t_1,y)-L(t_2,y)|\le \alpha_L|t_1-t_2|
\end{align}
\end{assumption}
% \subsection{Generalization Error of Macro PN Learning}
At the macro level, the average generalization error bound for all classes under normal circumstances is relatively easy to infer. However, due to the issue of extremely low abnormal class information in abnormal class samples, as much as $\frac{l-1}{l}$ of the information in the classification task samples can be redundant or even considered as noise. Therefore, the generalization error bound of coarse-grained PN (CGPN) learning that we derive is:
% while considering the proportion of redundant information 
\begin{theorem}
For any $\delta > 0$, with probability at least $1-\delta$:
% \begin{align}
%     &\hat{R}(g_{pn})-R(g^*)\nonumber\\\le& 4L_{l}\mathcal{R}_{|P_{macro}|,p^+_{1/l}}(\mathcal{G})+4L_{l}\mathcal{R}_{|N_{macro}|,p^-_{1/l}}(\mathcal{G})\nonumber\\+&\sqrt{\frac{2ln(4/\sigma)}{|P_{macro}|}}+\sqrt{\frac{2ln(4/\sigma)}{|N_{macro}|}}\nonumber\\+&Disc(\mathcal{G},p_1(x,y),p_{1/l}(x,y))
% \end{align}
% \begin{align}
%     \hat{R}(g_{pn})-R(g^*)\le &\frac{4L_{l}C_{\mathcal{G}}}{\sqrt{|P_{macro}|}}+\frac{4L_{l}C_{\mathcal{G}}}{\sqrt{|N_{macro}|}}\nonumber\\+&\sqrt{\frac{2ln(4/\delta)}{|P_{macro}|}}+\sqrt{\frac{2ln(4/\delta)}{|N_{macro}|}}\nonumber\\+&Disc(\mathcal{G},p_1(x,y),p_{1/l}(x,y))
% \end{align}
\begin{align}
    &\hat{R}(g_{cgpn})-R(g^*)\le\frac{2\cdot(\sigma_{macro}+1)\cdot\sqrt{\sigma_{micro+1}}\cdot \alpha_{L}\cdot C_{\mathcal{G}}}{\sigma_{macro}\cdot\sqrt{|P_{micro}|}}\nonumber\\&+\frac{\sigma_{macro}+1}{2\cdot\sigma_{macro}}\cdot\sqrt{\frac{2\cdot(\sigma_{micro}+1)\cdot ln(4/\delta)}{|P_{micro}|}}+\frac{2\cdot(\sigma_{macro}+1)
\cdot\sqrt{\sigma_{micro+1}}\cdot \alpha_L\cdot C_{\mathcal{G}}}{\sqrt{|U_{micro}|}}\nonumber\\&+\frac{\sigma_{macro}+1}{2}\cdot\sqrt{\frac{2\cdot (\sigma_{micro}+1)\cdot ln(4/\delta)}{|U_{micro}|}}+Disc(\mathcal{G},p_1(x,y),p_{1/l}(x,y))
\end{align}
where $g^* =\arg \min_{g\in\mathcal{G}}R(g)$
be the optimal decision function for $p_1(x,y)$ in $\mathcal{G}$, and
% \begin{align}
% &Disc(\mathcal{G},p_1(x,y),p_{1/l})\nonumber\\=&\max_{g\in\mathcal{G}}|p_{x,y\sim\ p_1(x,y)}(g(x)\not=y)-p_{x,y\sim\ p_{1/l}(x,y)}(g(x)\not=y)|
% \end{align}
\begin{align}
Disc(\mathcal{G},p_1(x,y),p_{1/l})=\max_{g\in\mathcal{G}}|p_{x,y\sim\ p_1(x,y)}(g(x)\not=y)-p_{x,y\sim\ p_{1/l}(x,y)}(g(x)\not=y)|
\end{align}
represents the discrepancy in distribution between the effective information proportion of $1/l$ and the effective information proportion of 1 for the function set $\mathcal{G}$. 
\end{theorem}
% \subsection{Generalization Error of MIL}
% 在经典MIL算法中，异常的bag中所有的instance都被视为异常，这会造成严重的偏差，其偏差程度由微观与宏观层面的不平衡程度$\sigma_{micro}$和$\sigma_{macro}$共同决定。
In classic MIL algorithms, all instances within an anomalous bag are treated as anomalous, which introduces significant bias. The extent of this bias is jointly determined by the imbalance levels at both the micro and macro levels, denoted as $\sigma_{micro}$ and $\sigma_{macro}$, respectively. Therefore, the generalization error bound of MIL that we derive is:
\begin{theorem}
For any $\delta > 0$, with probability at least $1-\delta$:
% \begin{spreadlines}[-1em]
\begin{align}
&\hat{R}(g_{mil})-R(g^*)\le \frac{2\cdot(\sigma_{macro}+1)\cdot \alpha_{L}\cdot C_{\mathcal{G}}}{\sigma_{macro}\cdot\sqrt{|P_{micro}|} }+\frac{2\cdot(\sigma_{macro}+1)
\cdot\alpha_{L}\cdot C_{\mathcal{G}}}{\sqrt{|U_{micro}|}}+\frac{\sigma_{macro}+1}{2\cdot\sigma_{macro}}\nonumber\\\cdot&\sqrt{\frac{2\cdot ln(4/\delta)}{|P_{micro}|}}+\frac{\sigma_{macro}+1}{2}\cdot\sqrt{\frac{2\cdot ln(4/\delta)}{|U_{micro}|}}+p_{inc}(\mathcal{G},g^*)+\frac{\sigma_{macro}+1}{2}\cdot\frac{\sigma_{micro}}{\sigma_{micro}+1}
\end{align}
% \end{spreadlines}
where $p_{inc}(\mathcal{G},g^*)=\max_{g\in\mathcal{G}}p(\arg\max_j g_{-1}(x_j)\not=\arg\max_j g^*_{-1}(x_j))$ means the inconsistency between the true closest abnormal micro-sample in the macro sample and the predicted one.
\end{theorem}
% \subsection{Generalization Error of Micro PU Learning}
Based on the derivation of the PU loss in \cref{bfgpu}, we found that directly optimizing unlabeled samples as abnormal samples essentially balances the loss of PN learning. Therefore, the generalization error bound of BFGPU that we derive is:
% Moreover, due to the absence of redundant information interference at the micro level and a larger sample size, the generalization error of fine-grained PU learning is much lower.
% The only concern is the additional error introduced during the transition from micro optimization goals to macro optimization goals, namely, the precision of the indicator function:
\begin{theorem}
For any $\delta > 0$, with probability at least $1-\delta$:
% \begin{align}
%     \hat{R}(g_{pu})-R(g^*)&\le \frac{4L_{l}C_{\mathcal{G}}}{\sqrt{|P_{micro}|}}+\frac{4L_{l}C_{\mathcal{G}}}{\sqrt{|U_{micro}|}}+\sqrt{\frac{2ln(4/\delta)}{|P_{micro}|}}\nonumber\\&+\sqrt{\frac{2ln(4/\delta)}{|U_{micro}|}}+p_{inc}(\mathcal{G},p_1(x,y))
% \end{align}
\begin{align}
\hat{R}(g_{bfgpu})-R(g^*)&\le\frac{4\cdot\alpha_{L}\cdot C_{\mathcal{G}}}{\sqrt{|P_{micro}|}}+\sqrt{\frac{2\cdot ln(4/\delta)}{|P_{micro}|}}+\frac{4\cdot \alpha_{L}\cdot C_{\mathcal{G}}}{\sqrt{|U_{micro}|}}+\sqrt{\frac{2\cdot ln(4/\delta)}{|U_{micro}|}}+p_{inc}(\mathcal{G},g^*)
%
% \end{align}
% p(j^*\not=\arg\min_j g_{-1}(x_j))
% \label{theorem2}
% where 
% \begin{align}
    % p_{inc}(\mathcal{G},p_1(x,y))&=\max_{g\in\mathcal{G}}\max_{\substack{x_i\in p_1(x)\\i\in [1,\dots,l]}}p(j^*\not=\arg\min_j g_{-1}(x_j))
\end{align}
% \begin{align}
% \end{align} 
\end{theorem}
The above theoretical results indicate that when the proportion of negative samples at the micro level is too low in the macro-level samples, the micro PU learning paradigm not only reduces the variance caused by model space complexity and confidence $\delta$ due to a large number of samples but also does not need to face the distribution discrepancy caused by a large amount of redundant or noisy information. BFGPU eliminates the bias in MIL, and unlike CGPN and MIL—which are heavily affected by $\sigma_{micro}$ and $\sigma_{macro}$—its error bound is not influenced by the class imbalance problem.
% The effectiveness of this paradigm also depends on the judgment of the micro samples closest to the abnormal class at the macro level.
\section{Conclusion}
% 我们讨论了现实应用场景下在异常信息含量极低的情况下完成文本异常检测的两难困境：在宏观上相似度过低，在微观上没有负类标注。我们将问题转化为了微观层面的不平衡PU学习问题，通过理论分析论证了方案的可行性，并提出了直接在微观层面优化宏观层面性能的PU学习目标，并基于此目标提出新的损失函数，并结合伪标注与检测阈值调整技术为这一问题提供了新的算法框架。实验验证了这一新的算法框架不仅可以解决微观上异常信息含量低的问题，更重要的是它在宏观层面负面文本含量低且微观层面负面信息在负面文本中含量低这种双维度极度不平衡问题上依然可以保持优异的性能。
We addressed the challenge of scarce and sparse anomalous detection in real-world scenarios: macroscopically, there is high similarity between normal and anomalous samples, and microscopically, labels are lacking and imbalanced. Accordingly, we propose a dual-imbalanced MIL problem. We transformed the problem into an imbalanced PU learning task at the micro level, demonstrating its feasibility through theoretical analysis. We proposed a solution that directly optimizes macro-level performance at the micro level using PU learning objectives and combined pseudo-labeling with threshold adjustment techniques to create a new framework. Experiments validated that this framework effectively tackles the issue of sparse negative information while maintaining strong performance even in extreme cases of imbalance at both macro and micro levels. 
% In future work, we will focus on more complex text detection tasks, such as extending the algorithm to dynamically handle unseen anomalies.
\newpage 
\section*{Ethics Statement}
We foresee no direct negative societal impacts resulting from this work, which is intended to advance research in computational music and creativity.
\section*{Reproducibility Statement}
We are fully committed to the reproducibility of our research. The code for this work has been open-sourced at https://github.com/BFGPU/BFGPU, and the experimental setup section of the paper provides a comprehensive description of all models and parameters used. We will subsequently focus on providing a more user-friendly interface to facilitate their use of the proposed method as much as possible.
% 这份工作的代码已经开源于https://github.com/BFGPU/BFGPU，且文章实验设置部分充分说明了所有模型和参数的使用。
\bibliography{iclr2026_conference}
\bibliographystyle{iclr2026_conference}
\appendix
\newpage
\section{LLMs Usage Statement}
We used Large Language Models (LLMs) to assist with the writing of this paper. Their primary role was to improve grammar, phrasing, and clarity.
\section{Algorithm Framework}
This section presents the specific algorithmic framework.
\begin{algorithm}[htbp]
   \caption{Balanced Fine-Grained PU Learning}
   \label{algorithm}
   \begin{algorithmic}
   \STATE{\bfseries Training Phase}
   \STATE {\bfseries Input:} macro positive dataset $P_{macro}$, macro negative dataset $N_{macro}$, the coefficient of $\hat{R}_{bfgpu}$ $\lambda_{bfgpu}$, the coefficient of $\hat{R}_{pse}$ $\lambda_{pse}$, learning rate $\eta$, the number of epochs $E$, class distribution prior $\pi$.
   \STATE {\bfseries Output:}  micro classifier $g$, threshold $T$.
   \STATE Split the macro-level data: $P_{micro}\leftarrow Split(P_{macro})$
   \STATE Split the macro-level data: $U_{micro}\leftarrow Split(N_{macro})$
   \STATE Initialize g with parameters $\theta$
   \FOR{$e=1$ {\bfseries to} $E$}
   \FOR{$P_{batch}, U_{batch}$ {\bfseries in} $P_{micro}, U_{micro}$}
   \STATE Get the probabilities $\hat{p}$ by \cref{hatp}
   \STATE Get the loss $\hat{R}_{bfgpu}$ by \cref{bfgpu}
   \STATE $\theta=\theta-\eta\nabla_\theta(\lambda_{bfgpu}\cdot \hat{R}_{bfgpu})$
   \ENDFOR
   \STATE Get $P_{pse}, N_{pse}$ by \cref{18,19}
   \FOR{$P_{batch}, N_{batch}$ {\bfseries in} $P_{pse}, N_{pse}$}
   \STATE Get the loss $\hat{R}_{pse}$ by \cref{20}
   \STATE $\theta=\theta-\eta\nabla_\theta(\lambda_{pse}\cdot \hat{R}_{pse})$
   \ENDFOR
   \ENDFOR
   \STATE Get the threshold $T$ by \cref{21} 
\end{algorithmic}
\begin{algorithmic}
    \STATE {\bfseries Testing Phase}
   \STATE {\bfseries Input:} macro test data $X_T$, micro classifier $g$, and threshold $T$.
   \STATE {\bfseries Output:} macro prediction $Y_T$.
   \FOR{$X_i$ {\bfseries in} $X_T$}
        % \STATE ${X_{micro}}_i\leftarrow Split({X_{macro}}_i)$
        \STATE Initial ${Y_T}_i\leftarrow +1$
        \FOR{$x_{ij}$ {\bfseries in} $X_i$}
        \IF{$g_{-1}(x_{ij})> T$}
        \STATE Update ${Y_T}_i\leftarrow -1$
        \ENDIF
        \ENDFOR
   \ENDFOR
\end{algorithmic}
\end{algorithm}
% \section*{The Pseudocode and Open-Source Information of the Algorithm}
% \label{opensource}
% For the sake of reproducibility, we have provided detailed algorithm pseudocode and released the PyTorch version of the code on GitHub. The open-source link is https://github.com/BFGPU/BFGPU.
% \section*{Limitations}
% \label{limitations}
% Due to the transition from macro to micro, the length of each text decreases while the total amount of text increases. Although the overall text length processed by the text model remains the same, micro approaches require more frequent loss calculations and gradient backpropagation compared to macro ones. Therefore, the micro algorithm generally has a longer runtime than the macro algorithm, but the performances of micro algorithms are significantly higher than those of macro ones.
\section{Proof of the Theories}
\label{proof}
\subsection{Proof of Theorem 4.1}
Consider directly learning from macro-level data at the macro level, where the proportion of effective information is $1/l$ in the macro, and then testing on data with the same proportion. It can be shown that there exists $\delta > 0$, with at least a probability of $1 - \delta$:
\begin{align}
    &\hat{R}(g_{cgpn})-R(g_{1/l}^*)\nonumber\\\le& 
    \frac{1}{2}\cdot(\frac{4\cdot \alpha_{L}\cdot C_{\mathcal{G}}}{\sqrt{|P_{micro}|/(\sigma_{micro}+1)}}+\sqrt{\frac{2\cdot ln(4/\delta)}{|P_{micro}|/(\sigma_{micro}+1)}})/\frac{\sigma_{macro}}{\sigma_{macro}+1}\nonumber\\+&\frac{1}{2}\cdot(\frac{4\cdot \alpha_{L}\cdot C_{\mathcal{G}}}{\sqrt{|U_{micro}|/(\sigma_{micro}+1)}}+\sqrt{\frac{2\cdot ln(4/\delta)}{|U_{micro}|/(\sigma_{micro}+1)}})/\frac{1}{\sigma_{macro}+1}
\end{align}
Then, considering the data distribution caused by redundant information differs from the data distribution when there is no redundant information, it is not difficult to conclude that:
\begin{align}
    R(g_{1/l}^*)-R_(g^*)\le Disc(\mathcal{G},p_1(x,y),p_{1/l}(x,y))
\end{align}
By combining the two equations, we can measure the gap between the macro PN error and the error of the optimal classifier at the macro level when there is no redundant information. Ultimately proved for any $\delta > 0$, with probability at least $1-\delta$:
\begin{align}
    &\hat{R}(g_{cgpn})-R(g^*)\nonumber\\=&(\hat{R}(g_{cgpn})-R(g_{1/l}^*))+(R(g_{1/l}^*)-R_(g^*))\nonumber\\\le&\frac{2\cdot(\sigma_{macro}+1)\cdot\sqrt{\sigma_{micro+1}}\cdot \alpha_{L}\cdot C_{\mathcal{G}}}{\sigma_{macro}\cdot\sqrt{|P_{micro}|}}\nonumber\\+&\frac{\sigma_{macro}+1}{2\cdot\sigma_{macro}}\cdot\sqrt{\frac{2\cdot(\sigma_{micro}+1)\cdot ln(4/\delta)}{|P_{micro}|}}+\frac{2\cdot(\sigma_{macro}+1)
\cdot\sqrt{\sigma_{micro+1}}\cdot \alpha_L\cdot C_{\mathcal{G}}}{\sqrt{|U_{micro}|}}\nonumber\\+&\frac{\sigma_{macro}+1}{2}\cdot\sqrt{\frac{2\cdot (\sigma_{micro}+1)\cdot ln(4/\delta)}{|U_{micro}|}}+Disc(\mathcal{G},p_1(x,y),p_{1/l}(x,y))
\end{align}
\subsection{Proof of Theorem 4.2}
% 考虑将所有微观层面的样本都使用宏观样本的标注，则对于正常样本，用于训练的标注全部正确，而对于异常样本，标注误差的上界为$\frac{\sigma_{micro}}{\sigma_{micro}+1}$。在宏观层面的不平衡比例为$\sigma_{macro}$时，不难得出使用MIL方法进行训练的平衡误差上界为
Consider assigning the macro-level labels to all micro-level samples. For normal samples, all training labels are correct; whereas for anomalous samples, the upper bound of labeling error is $\frac{\sigma_{micro}}{\sigma_{micro}+1}$. When the class imbalance ratio at the macro level is $\sigma_{macro}$, it is straightforward to derive the balanced error upper bound for training with the MIL method. It can be shown that there exists $\delta > 0$, with at least a probability of $1 - \delta$:

\begin{align}
    &\hat{R}(g_{mil})-R(g_{micro}^*)\nonumber\\\le&\frac{1}{2}\cdot( \frac{4\cdot\alpha_{L}\cdot C_{\mathcal{G}}}{\sqrt{|P_{micro}|}}+\sqrt{\frac{2\cdot ln(4/\delta)}{|P_{micro}|}})/\frac{\sigma_{macro}}{\sigma_{macro}+1}\nonumber\\+&\frac{1}{2}\cdot(\frac{4\cdot L_{l}\cdot C_{\mathcal{G}}}{\sqrt{|U_{micro}|}}+\sqrt{\frac{2\cdot ln(4/\delta)}{|U_{micro}|}}+\frac{\sigma_{micro}}{\sigma_{micro}+1})/\frac{1}{\sigma_{macro}+1}
\end{align}

Considering the inconsistency between micro-level optimization goals and macro-level optimization goals, it is not difficult to conclude that:
\begin{align}
    R(g_{micro}^*)-R(g^*)\le p_{inc}(\mathcal{G},g^*)
\end{align}
By combining the two equations, we can measure the gap between the MIL error and the error of the optimal classifier at the macro level. Ultimately proved for any $\delta > 0$, with probability at least $1-\delta$:
For any $\delta > 0$, with probability at least $1-\delta$:

\begin{align}
    &\hat{R}(g_{mil})-R(g^*)\nonumber\\=&(\hat{R}(g_{mil})-R(g_{micro}^*))+(R(g_{micro}^*)-R(g^*))\nonumber\\\le& \frac{2\cdot(\sigma_{macro}+1)\cdot \alpha_{L}\cdot C_{\mathcal{G}}}{\sigma_{macro}\cdot\sqrt{|P_{micro}|} }+\frac{2\cdot(\sigma_{macro}+1)
\cdot\alpha_{L}\cdot C_{\mathcal{G}}}{\sqrt{|U_{micro}|}}+\frac{\sigma_{macro}+1}{2\cdot\sigma_{macro}}\cdot\sqrt{\frac{2\cdot ln(4/\delta)}{|P_{micro}|}}\nonumber\\+&\frac{\sigma_{macro}+1}{2}\cdot\sqrt{\frac{2\cdot ln(4/\delta)}{|U_{micro}|}}+p_{inc}(\mathcal{G},g^*)+\frac{\sigma_{macro}+1}{2}\cdot\frac{\sigma_{micro}}{\sigma_{micro}+1}
\end{align}
\subsection{Proof of Theorem 4.3}
By directly optimizing the balanced PU loss at the micro level, we can achieve unbiased optimization of macro-level performance. Taking into account the error introduced by converting the macro-level problem to the micro level, we ultimately obtain an error upper bound for BFGPU that is independent of both imbalance ratios $\sigma_{macro}$ and $\sigma_{micro}$. It can be shown that there exists $\delta > 0$, with at least a probability of $1 - \delta$:

\begin{align}
    &\hat{R}(g_{bfgpu})-R(g_{micro}^*)\nonumber\\\le& \frac{4\cdot L_{l}\cdot C_{\mathcal{G}}}{\sqrt{|P_{micro}|}}+\frac{4\cdot L_{l}\cdot C_{\mathcal{G}}}{\sqrt{|U_{micro}|}}\nonumber\\+&\sqrt{\frac{2\cdot ln(4/\delta)}{|P_{micro}|}}+\sqrt{\frac{2\cdot ln(4/\delta)}{|U_{micro}|}}
\end{align}
Considering the inconsistency between micro-level optimization goals and macro-level optimization goals, it is not difficult to conclude that:
\begin{align}
    R(g_{micro}^*)-R(g^*)\le p_{inc}(\mathcal{G},g^*)
\end{align}
By combining the two equations, we can measure the gap between the micro PU error and the error of the optimal classifier in the marco level when there is no redundant information. Ultimately proved for any $\delta > 0$, with probability at least $1-\delta$:
For any $\delta > 0$, with probability at least $1-\delta$:

\begin{align}
    &\hat{R}(g_{bfgpu})-R(g^*)\nonumber\\=&(\hat{R}(g_{bfgpu})-R(g_{micro}^*))+(R(g_{micro}^*)-R(g^*))\nonumber\\\le& \frac{4\cdot L_{l}\cdot C_{\mathcal{G}}}{\sqrt{|P_{micro}|}}+\frac{4\cdot L_{l}\cdot C_{\mathcal{G}}}{\sqrt{|U_{micro}|}}\nonumber\\+&\sqrt{\frac{2\cdot ln(4/\delta)}{|P_{micro}|}}+\sqrt{\frac{2\cdot ln(4/\delta)}{|U_{micro}|}}+p_{inc}(\mathcal{G},g^*)
\end{align}
% \section{Sensitivity Analysis}
% We have supplemented the sensitivity analysis of hyperparameters $\lambda_{bfgpu}$ and $\lambda_{pse}$. Experiments have demonstrated that the performance of the algorithm remains relatively stable under different hyperparameter settings, so there is no need to worry too much about hyperparameter tuning.

\section{Additional Experimental Results}
It is evident that our proposed BFGPU achieves optimal performance in most settings. In only a few cases does the nnPU or robustPU loss function achieve optimal performance, but this also depends on our proposed ADT technique.
\subsection{Experiments with Synthetic Datasets}
In addition to the SST-2 dataset, we conducted experiments on Sentiment140 which is a large short-text dataset with a focus on high imbalance ratios, setting $\sigma_{micro}$ to [2, 4, 6, 8, 10].

Due to space limitations in the main text, the additional experiments on the SST-2 and Sentiment140 datasets are provided in \cref{sst2_f1_table,sentiment_table} in this appendix.
\begin{table*}[htbp]
\centering
\small
\caption{The table presents the F1 scores of various algorithms under varying values of $\sigma_{\text{micro}} \in \{2, 4, 6, 8, 10\}$, along with the estimated AUC with respect to changes in F1 on the SST-2 Dataset.}
\label{sst2_f1_table}
% \vskip 0.15in
\resizebox{1\textwidth}{!}{\begin{tabular}{c c c c c c c c}
    \hline\hline
&\multirow{2}{*}{Methods} & \multicolumn{5}{c}{F1 Score} & \multirow{2}{*}{$AUC_{F1}$}\\\cline{3-7}
&&2&4&6&8&10&\\
    \hline\hline
    \multirow{2}{*}{Macro}& Supervised&$83.63\pm0.21$&$79.35\pm1.04$&$78.15\pm2.81$&$75.77\pm5.79$&$75.50\pm3.23$&$78.48$\\
    % Coarse-Grained Sad&$78.19\pm0.27$&$69.73\pm5.38$&$68.54\pm4.37$&$46.10\pm6.41$&$36.38\pm19.77$\\
    &DeepSAD&$75.24\pm1.86$&$47.14\pm13.51$&$28.13\pm9.66$&$13.04\pm9.22$&$7.00\pm7.06$&$34.11$\\\hline
    \multirow{5}{*}{MIL}
    &MaxPooling&$85.28\pm7.74$&$52.34\pm4.00$&$37.42\pm3.35$&$12.38\pm4.36$&$19.49\pm4.03$&$41.38$\\
    &TopkPooling&$-$&$69.95\pm6.85$&$70.39\pm4.89$&$61.75\pm4.28$&$44.44\pm38.49$&$61.63$\\
    &Attention&$80.54\pm0.51$&$62.39\pm7.22$&$45.00\pm36.50$&$0.00\pm0.00$&$48.79\pm30.96$&$47.34$\\
    &FGSA&$49.38\pm43.89$&$44.44\pm38.49$&$66.67\pm0.00$&$44.44\pm38.49$&$44.44\pm38.49$&$49.87$\\
    &PReNET&$73.67\pm10.48$&$53.84\pm7.58$&$48.68\pm6.08$&$48.77\pm16.00$&$36.10\pm12.41$&$52.21$\\\hline
    \multirow{7}{*}{Micro}&DeepSAD&$51.95\pm12.30$&$14.95\pm2.18$&$9.01\pm7.18$&$9.92\pm2.40$&$14.18\pm4.29$&$20.00$\\
    &DeepSVDD&$24.98\pm20.38$&$12.08\pm5.99$&$26.98\pm19.65$&$10.96\pm1.51$&$13.30\pm4.16$&$17.66$\\&uPU&$76.82\pm3.40$&$64.43\pm6.62$&$62.13\pm6.88$&$60.79\pm10.15$&$67.56\pm1.27$&$66.35$\\ 
    &nnPU&$84.00\pm1.25$&$66.74\pm0.10$&$66.67\pm0.00$&$66.67\pm0.00$&$66.67\pm0.00$&$70.15$\\
    &balancedPU&$80.75\pm4.89$&$78.73\pm6.27$&$52.51\pm37.17$&$61.83\pm7.17$&$70.48\pm5.39$&$68.86$\\
    &robustPU&$86.31\pm0.72$&$76.10\pm6.57$&$74.79\pm3.74$&$63.29\pm10.26$&$62.90\pm6.51$&$72.68$\\
    % &uPU+ADT&$84.09\pm1.08$&$77.99\pm1.55$&$75.59\pm2.66$&$69.91\pm1.50$&$65.42\pm9.19$&$74.60$\\
    % &nnPU+ADT&$87.75\pm0.60$&\bm{$85.13\pm0.84$}&$81.56\pm1.00$&$78.85\pm2.79$&$78.39\pm2.57$&$82.34$\\
    % &balancedPU+ADT&$80.94\pm6.40$&$79.54\pm4.81$&$57.02\pm32.67$&$65.03\pm13.16$&$76.11\pm1.94$&$71.73$\\
    % &robustPU+ADT&$86.91\pm1.48$&$59.86\pm45.37$&$80.23\pm2.51$&$74.05\pm8.97$&$48.39\pm42.41$&69.89\\
    &BFGPU&\bm{$88.73\pm0.61$}&\bm{$84.13\pm0.30$}&\bm{$83.96\pm1.10$}&\bm{$81.95\pm1.87$}&\bm{$84.41\pm1.64$}&\bm{$84.64$}\\
    \hline\hline
\end{tabular}}
% \vskip 0.01in
\end{table*}
\begin{table}[htbp]
\centering
\small
\caption{The table presents AvgAcc and F1 scores of various algorithms under varying values of $\sigma_{\text{micro}} \in \{2, 4, 6,8,10\}$, along with the estimated AUC with respect to changes in AvgAcc and F1 on the Sentiment140 Dataset.}
\label{sentiment_table}
% \vskip 0.15in
\resizebox{1\textwidth}{!}{
\begin{tabular}{c c c c c c c c}
\hline\hline
&\multirow{2}{*}{Method} &\multicolumn{5}{c}{AvgAcc} &\multirow{2}{*}{$AUC_{AvgAcc}$}\\\cline{3-7}
&&2&4&6&8&10&\\
    \hline\hline
    \multirow{2}{*}{Macro}& Supervised&$74.37\pm0.40$&$69.80\pm0.37$&$66.81\pm0.48$&$65.09\pm0.19$&$62.20\pm0.48$&$67.65$\\
    % Coarse-Grained AD&$75.95\pm1.52$&$73.60\pm0.26$&$70.58\pm0.48$&$68.61\pm0.22$&$61.41\pm6.18$\\
    &DeepSAD&$57.03\pm4.17$&$53.41\pm0.94$&$52.56\pm1.24$&$51.37\pm0.63$&$51.21\pm0.99$&$53.12$\\\hline
    \multirow{5}{*}{MIL}
    &MaxPooling&$60.79\pm2.63$&$62.51\pm9.54$&$56.78\pm5.84$&$49.95\pm0.09$&$52.03\pm3.41$&$56.41$\\
    &TopkPooling&$-$&$65.08\pm1.90$&$62.14\pm2.28$&$56.31\pm5.96$&$52.95\pm5.07$&$59.12$\\
    &Attention&$71.27\pm0.36$&$64.97\pm3.24$&$52.88\pm4.98$&$50.06\pm0.10$&$50.16\pm0.28$&$57.87$\\
    &FGSA&$64.17\pm3.13$&$53.46\pm4.03$&$50.00\pm0.00$&$50.00\pm0.00$&$50.00\pm0.00$&$53.53$\\
    &PReNET&$70.41\pm3.74$&$57.11\pm7.11$&$55.56\pm4.83$&$54.85\pm2.93$&$50.00\pm0.00$&$57.59$\\\hline
    \multirow{7}{*}{Micro}&DeepSAD&$51.07\pm2.14$&$51.80\pm2.05$&$65.23\pm2.04$&$63.99\pm8.28$&$58.81\pm8.64$&$58.18$\\
    &DeepSVDD&$49.65\pm0.55$&$50.09\pm0.18$&$49.70\pm0.58$&$49.83\pm0.72$&$50.27\pm0.29$&$49.91$\\&uPU&$68.81\pm1.02$&$62.73\pm2.48$&$61.23\pm3.25$&$57.54\pm5.37$&$54.31\pm3.33$&$60.92$\\
    &nnPU&$69.66\pm1.01$&$50.00\pm0.00$&$50.00\pm0.00$&$50.00\pm0.00$&$50.00\pm0.00$&$53.93$\\
    &balancedPU&$72.55\pm2.23$&$70.26\pm0.16$&$66.16\pm1.53$&$50.00\pm0.00$&$50.68\pm0.54$&$61.93$\\
    &robustPU&$74.65\pm0.70$&$70.13\pm1.63$&$66.94\pm0.40$&$59.14\pm8.58$&$62.88\pm0.27$&$66.74$\\
    % &uPU+ADT&$72.53\pm0.09$&$67.37\pm0.74$&$65.41\pm1.04$&$61.25\pm1.11$&$59.29\pm0.61$&$65.17$\\
    % &nnPU+ADT&$74.64\pm0.28$&$70.34\pm0.68$&$67.64\pm0.39$&$64.40\pm0.69$&\bm{$64.87\pm0.68$}&$68.38$\\
    % &balancedPU+ADT&$72.98\pm0.61$&$68.21\pm0.91$&$67.17\pm1.15$&$59.71\pm3.54$&$56.17\pm6.68$&$64.83$\\
    % &robustPU+ADT&$75.17\pm0.48$&$70.98\pm2.09$&$69.04\pm0.89$&$60.03\pm5.42$&$65.49\pm1.62$&$68.14$\\
    &BFGPU&\bm{$75.86\pm1.01$}&\bm{$70.68\pm0.97$}&\bm{$67.95\pm1.53$}&\bm{$66.34\pm1.25$}&\bm{$64.45\pm0.35$}&\bm{$69.06$}\\
    \hline\hline

&\multirow{2}{*}{Method} &\multicolumn{5}{c}{F1 Score} &\multirow{2}{*}{$AUC_{F1}$}\\\cline{3-7}
&&2&4&6&8&10&\\
    \hline\hline
    \multirow{2}{*}{Macro}& Supervised&$73.99\pm0.72$&$68.75\pm0.44$&$64.94\pm1.66$&$63.86\pm0.52$&$59.50\pm1.48$&$66.21$\\
    % Coarse-Grained AD&$75.54\pm1.54$&$73.11\pm0.25$&$70.16\pm0.61$&$68.05\pm0.13$&$60.93\pm6.09$\\
    &DeepSAD&$57.24\pm4.26$&$53.61\pm1.16$&$52.03\pm1.33$&$51.55\pm0.71$&$51.83\pm0.77$&$53.25$\\\hline
    \multirow{5}{*}{MIL}
    &MaxPooling&$37.85\pm7.23$&$43.23\pm30.95$&$27.71\pm23.78$&$21.92\pm37.96$&$9.96\pm16.53$&$28.13$\\
    &TopkPooling&$-$&$52.83\pm4.60$&$47.32\pm4.56$&$26.12\pm23.20$&$34.85\pm33.37$&$40.28$\\
    &Attention&$67.64\pm0.52$&$56.56\pm5.58$&$37.40\pm34.07$&$22.85\pm37.96$&$23.34\pm37.56$&$41.56$\\
    &FGSA&$52.19\pm8.97$&$16.28\pm17.59$&$44.44\pm38.49$&$44.44\pm38.49$&$44.44\pm38.49$&$40.36$\\

    &PreNET&$63.36\pm8.65$&$26.89\pm24.61$&$44.62\pm19.17$&$20.65\pm12.24$&$66.67\pm0.00$&$44.44$\\\hline
    \multirow{7}{*}{Micro}&DeepSAD&$54.65\pm1.80$&$55.89\pm1.76$&$67.50\pm1.80$&$66.55\pm7.42$&$61.86\pm7.61$&$61.29$\\
    &DeepSVDD&$53.33\pm0.67$&$54.27\pm0.20$&$53.37\pm0.71$&$53.61\pm1.05$&$54.11\pm0.49$&$53.74$\\
    &uPU&$60.88\pm2.42$&$57.32\pm10.29$&$48.65\pm13.84$&$60.66\pm4.54$&$66.66\pm0.21$&$58.83$\\
    &nnPU&$75.50\pm0.56$&$66.67\pm0.00$&$66.67\pm0.00$&$66.67\pm0.00$&$66.67\pm0.00$&$68.44$\\
    &balancedPU&$72.00\pm1.80$&$69.98\pm0.95$&$70.80\pm0.84$&$44.44\pm31.43$&\bm{$66.83\pm0.24$}&$64.81$\\
    &robustPU&$73.43\pm1.05$&$67.66\pm1.83$&$61.80\pm2.36$&$64.00\pm2.35$&$55.89\pm0.48$&$$64.55$$\\
    % &uPU+ADT&$73.81\pm0.08$&$69.25\pm0.67$&$66.25\pm1.23$&$60.57\pm1.56$&$57.65\pm0.71$&$65.51$\\
    % &nnPU+ADT&$75.71\pm0.17$&$72.46\pm0.61$&$69.93\pm0.10$&$66.34\pm0.36$&\bm{$67.27\pm1.10$}&$70.34$\\
    % &balancedPU+ADT&$72.93\pm0.62$&$68.81\pm1.89$&$68.78\pm0.98$&$58.49\pm6.25$&$53.03\pm12.28$&$64.41$\\
    % &robustPU+ADT&$75.34\pm0.69$&$71.46\pm2.84$&$70.10\pm0.81$&$57.85\pm8.03$&$67.04\pm2.15$&$68.36$\\
    &BFGPU&\bm{$77.21\pm0.73$}&\bm{$73.27\pm0.95$}&\bm{$69.80\pm3.18$}&\bm{$67.21\pm2.94$}&$66.40\pm0.68$&$\bm{70.78}$\\
    \hline\hline
\end{tabular}}
% \vskip -0.15in
\end{table}
We present a comparative experiment on two long text datasets: IMDB and Amazon. The parameter $\sigma_{micro}$ is set to [2, 3, 4, 5]. Due to space limitations in the main text, the additional experiments on the datasets Amazon are shown in \cref{imdb_f1_table,amaozn_table} in this appendix.
\begin{table*}[htbp]
\centering
\small
\caption{The table presents the AvgAcc and F1 scores of various algorithms under varying values of $\sigma_{\text{micro}} \in \{2, 3, 4, 5\}$, along with the estimated AUC on the IMDB Dataset.}
\label{imdb_f1_table}
% \vskip 0.05in
\begin{tabular}{c c c c c c c}
\hline\hline
&\multirow{2}{*}{Method} &\multicolumn{4}{c}{AvgAcc} & \multirow{2}{*}{$AUC_{AvgAcc}$}\\\cline{3-6}
&&2&3&4&5&\\
    \hline\hline
    \multirow{2}{*}{Macro}& Supervised&$76.75\pm1.45$&$70.62\pm0.83$&$62.61\pm4.70$&$58.33\pm4.81$&$67.08$\\
    % Coarse-Grained AD&$57.74\pm0.17$&$52.99\pm1.93$&$52.86\pm0.92$&$51.72\pm0.92$\\
    &DeepSAD&$52.17\pm0.56$&$52.57\pm1.20$&$55.01\pm1.05$&$50.76\pm0.77$&$52.63$\\\hline
    \multirow{5}{*}{MIL} 
    &MaxPooling&$85.76\pm2.55$&$74.30\pm18.50$&$60.31\pm15.71$&$55.56\pm9.63$&$68.98$\\
    &TopkPooling&$-$&$82.23\pm1.22$&$83.27\pm0.61$&$82.41\pm2.17$&$82.63$\\
    &Attention&$82.58\pm7.37$&$85.98\pm1.32$&$80.66\pm3.81$&$78.62\pm1.54$&$81.96$\\
    &FGSA&$82.67\pm1.60$&$63.60\pm13.43$&$60.49\pm14.59$&$58.64\pm14.97$&$66.35$\\
    &PReNET&$84.53\pm1.50$&$80.54\pm1.50$&$71.78\pm4.97$&$73.56\pm2.49$&$77.60$\\\hline
    \multirow{7}{*}{Micro}&DeepSAD&$62.86\pm4.33$&$60.00\pm1.43$&$57.80\pm1.46$&$56.84\pm0.12$&$59.38$\\
    &DeepSVDD&$52.70\pm0.61$&$51.73\pm1.13$&$51.96\pm0.09$&$51.76\pm2.58$&$52.04$\\
    &uPU&$82.93\pm1.14$&$76.30\pm1.51$&$78.57\pm5.37$&$72.29\pm3.30$&$77.52$\\
    &nnPU&$84.53\pm1.49$&$58.39\pm5.95$&$50.00\pm0.00$&$50.00\pm0.00$&$60.73$\\
    &balancedPU&$85.78\pm1.39$&$77.39\pm3.90$&$84.85\pm1.14$&$78.77\pm4.17$&$81.70$\\
    &robustPU&$87.82\pm0.37$&$86.63\pm1.11$&$85.49\pm0.45$&$72.58\pm19.57$&$83.13$\\
    % &uPU+ADT&$85.84\pm1.14$&$82.35\pm1.14$&$83.39\pm1.18$&$80.97\pm1.01$&$83.14$\\
    % &nnPU+ADT&$87.93\pm0.27$&$86.03\pm0.67$&$85.47\pm1.60$&\bm{$83.99\pm0.55$}&$85.86$\\
    % &balancedPU+ADT&$85.99\pm1.22$&$78.09\pm4.24$&$85.03\pm8.39$&$79.86\pm4.15$&$82.24$\\
    % &robustPU+ADT&$87.96\pm0.93$&$86.37\pm1.23$&$85.60\pm0.18$&$69.13\pm15.83$&$82.27$\\
    &BFGPU&\bm{$88.13\pm1.07$}&\bm{$87.93\pm0.58$}&\bm{$86.02\pm0.69$}&\bm{$83.64\pm0.78$}&\bm{$86.43$}\\
\hline\hline
&\multirow{2}{*}{Method} &\multicolumn{4}{c}{F1 Score} & \multirow{2}{*}{$AUC_{F1}$}\\\cline{3-6}
&&2&3&4&5\\
    \hline\hline
    \multirow{2}{*}{Macro}& Supervised&$77.73\pm2.22$&$73.77\pm0.92$&$63.98\pm10.18$&$50.24\pm22.06$&$66.43$\\
    % Coarse-Grained AD&$57.84\pm0.64$&$53.10\pm1.89$&$52.42\pm0.73$&$51.61\pm1.29$\\
    &DeepSAD&$52.22\pm0.59$&$52.91\pm1.37$&$54.84\pm1.33$&$50.82\pm1.21$&$52.70$\\\hline
    \multirow{5}{*}{MIL}
    &MaxPooling&$84.89\pm3.51$&$59.75\pm41.13$&$29.08\pm41.11$&$20.76\pm35.96$&48.62\\
    &TopkPooling&$-$&$79.15\pm1.95$&$81.37\pm1.30$&$80.26\pm2.95$&$80.26$\\
    &Attention&$82.27\pm7.82$&$85.67\pm1.41$&$78.78\pm4.93$&$75.43\pm2.50$&$80.54$\\
    &FGSA&$81.50\pm3.14$&$39.66\pm36.71$&$52.30\pm30.53$&$46.00\pm39.91$&$54.87$\\
    &PReNET&$82.83\pm2.11$&$77.07\pm2.40$&$61.39\pm9.70$&$65.83\pm4.80$&$71.78$\\\hline
    \multirow{7}{*}{Micro}&DeepSAD&$65.45\pm3.58$&$62.67\pm1.56$&$60.51\pm1.16$&$59.78\pm0.85$&$62.10$\\
    &DeepSVDD&$55.45\pm1.36$&$55.99\pm1.44$&$55.99\pm0.60$&$55.14\pm2.11$&$55.64$\\
    &uPU&$80.95\pm2.18$&$71.34\pm2.77$&$74.05\pm8.51$&$63.94\pm6.10$&$72.57$\\
    &nnPU&$86.15\pm1.03$&$70.68\pm2.85$&$66.67\pm0.00$&$66.67\pm0.00$&$72.54$\\
    &balancedPU&$85.94\pm1.78$&$79.12\pm3.57$&$85.69\pm0.93$&$80.93\pm3.70$&$82.92$\\
    &robustPU&$87.52\pm0.53$&$86.24\pm1.29$&$85.06\pm0.79$&$77.80\pm9.68$&$84.15$\\
    % &uPU+ADT&$86.37\pm1.12$&$83.40\pm1.02$&$84.62\pm1.06$&$82.44\pm0.74$&$84.21$\\
    % &nnPU+ADT&$88.10\pm0.37$&$86.48\pm0.57$&$86.11\pm1.43$&\bm{$84.86\pm0.30$}&$86.39$\\
    % &balancedPU+ADT&$86.04\pm1.30$&$78.02\pm4.88$&$85.66\pm0.76$&$80.33\pm4.84$&$82.51$\\
    % &robustPU+ADT&\bm{$88.24\pm0.91$}&$86.73\pm1.33$&$86.16\pm0.15$&$65.79\pm19.75$&$81.73$\\
    &BFGPU&\bm{$88.12\pm1.24$}&\bm{$88.36\pm0.58$}&\bm{$86.57\pm0.69$}&\bm{$84.61\pm0.43$}&\bm{$86.92$}\\
    \hline\hline
\end{tabular}
% \vskip -0.15in
\end{table*}
\begin{table*}[htbp]
\centering
\small
\caption{The table presents AvgAcc and F1 scores of various algorithms under varying values of $\sigma_{\text{micro}} \in \{2, 3, 4, 5\}$, along with the estimated AUC with respect to changes in AvgAcc and F1 on the Amazon Dataset.}
\label{amaozn_table}
% \vskip 0.15in
\begin{tabular}{c c c c c c c}
\hline\hline
&\multirow{2}{*}{Method} &\multicolumn{4}{c}{AvgAcc} & \multirow{2}{*}{$AUC_{AvgAcc}$}\\\cline{3-6}
&&2&3&4&5\\
    \hline\hline
    \multirow{2}{*}{Macro}& Supervised&$89.11\pm0.19$&$87.82\pm0.23$&$84.74\pm0.95$&$81.97\pm0.60$&$85.91$\\
    % Coarse-Grained AD&$88.35\pm0.23$&$85.99\pm0.21$&$83.33\pm0.55$&$78.94\pm0.79$\\
    &DeepSAD&$85.55\pm0.43$&$71.88\pm6.80$&$63.31\pm1.65$&$60.83\pm3.43$&$70.39$\\\hline
    \multirow{5}{*}{MIL}
    &MaxPooling&$87.09\pm4.28$&$85.25\pm3.01$&$82.38\pm1.18$&$78.96\pm4.81$&$83.42$\\
    &TopkPooling&$-$&$86.15\pm0.11$&$84.12\pm0.89$&$82.59\pm1.74$&$84.27$\\&Attention&$87.96\pm0.93$&$86.37\pm1.23$&$85.60\pm0.18$&$69.13\pm15.83$&$82.27$\\
    &FGSA&$59.46\pm16.39$&$64.73\pm14.45$&$52.50\pm4.33$&$50.00\pm0.00$&$56.67$\\ 
    &PReNET&$87.15\pm0.68$&$84.00\pm2.50$&$77.88\pm2.86$&$79.40\pm1.56$&$82.11$\\\hline
    \multirow{7}{*}{Micro}&DeepSAD&$62.86\pm9.94$&$52.14\pm0.99$&$51.20\pm2.28$&$50.84\pm0.17$&$54.26$\\
    &DeepSVDD&$50.2\pm1.36$&$51.19\pm0.74$&$50.45\pm1.06$&$50.44\pm0.59$&$50.60$\\
    &uPU&$82.48\pm0.38$&$76.92\pm2.66$&$74.15\pm2.31$&$74.69\pm0.18$&$77.06$\\
    &nnPU&$85.11\pm1.40$&$74.02\pm5.18$&$50.00\pm0.00$&$50.00\pm0.00$&$64.78$\\
    &balancedPU&$80.59\pm7.70$&$81.69\pm5.52$&$80.84\pm6.93$&$82.91\pm8.14$&$81.51$\\
    &robustPU&$89.28\pm0.52$&$87.07\pm0.63$&$86.34\pm0.34$&$82.47\pm1.08$&$86.29$\\
    % &uPU+ADT&$86.71\pm0.56$&$84.22\pm0.26$&$80.56\pm0.73$&$80.35\pm0.15$&$82.96$\\
    % &nnPU+ADT&$88.61\pm0.49$&$88.22\pm0.21$&$86.09\pm0.45$&$84.82\pm0.95$&$86.94$\\
    % &balancedPU+ADT&$82.13\pm7.87$&$81.69\pm5.49$&$80.59\pm7.00$&$82.91\pm0.93$&$81.83$\\
    % &robustPU+ADT&$89.58\pm0.18$&$80.27\pm11.09$&$86.10\pm0.95$&$75.50\pm18.14$&$82.86$\\
    &BFGPU&\bm{$89.87\pm0.13$}&\bm{$88.15\pm0.97$}&\bm{$86.71\pm0.18$}&\bm{$85.57\pm0.67$}&\bm{$87.58$}\\
    \hline\hline

&\multirow{2}{*}{Method} &\multicolumn{4}{c}{F1 Score} & \multirow{2}{*}{$AUC_{F1}$}\\\cline{3-6}
&&2&3&4&5&\\
    \hline\hline
    \multirow{2}{*}{Macro}& Supervised&$89.10\pm0.20$&$87.86\pm0.26$&$84.73\pm0.95$&$82.15\pm0.51$&$85.96$\\
    &DeepSAD&$88.22\pm0.25$&$85.87\pm0.15$&$83.02\pm0.60$&$78.63\pm0.93$&$83.94$\\\hline
    \multirow{5}{*}{MIL}
    &MaxPooling&$85.81\pm6.10$&$82.82\pm4.57$&$79.60\pm1.87$&$74.56\pm8.00$&$80.70$\\
    &TopkPooling&$-$&$84.80\pm0.17$&$82.30\pm1.31$&$80.26\pm2.73$&$82.45$\\
    &Attention&$88.24\pm0.91$&$86.73\pm1.33$&$86.16\pm0.15$&$65.79\pm19.75$&$81.73$\\
    &FGSA&$47.01\pm20.89$&$63.63\pm14.12$&$10.08\pm17.43$&$44.44\pm38.49$&$41.29$\\
    &PReNET&$86.10\pm0.88$&$81.97\pm3.58$&$72.85\pm4.66$&$75.66\pm2.73$&$79.14$\\\hline
    \multirow{7}{*}{Micro}&DeepSAD&$85.47\pm0.54$&$72.34\pm6.42$&$63.52\pm1.48$&$59.47\pm3.18$&$70.20$\\
    &DeepSVDD&$47.01\pm40.89$&$63.63\pm14.12$&$10.08\pm17.43$&$44.44\pm38.49$&$41.29$\\
    &uPU&$79.82\pm0.62$&$71.43\pm4.48$&$67.49\pm4.27$&$68.79\pm3.65$&$71.88$\\
    &nnPU&$86.63\pm1.09$&$79.34\pm3.11$&$66.67\pm0.00$&$66.67\pm0.00$&$74.83$\\
    &balancedPU&$78.50\pm11.50$&$81.24\pm6.09$&$81.38\pm6.31$&$83.76\pm0.71$&$81.22$\\
    &robustPU&$89.06\pm0.71$&$86.60\pm0.83$&$85.85\pm0.70$&$82.18\pm1.25$&$85.92$\\
    % &uPU+ADT&$87.33\pm0.51$&$85.01\pm0.39$&$81.59\pm0.91$&$81.49\pm0.16$&$83.86$\\
    % &nnPU+ADT&$88.79\pm0.52$&$88.64\pm0.23$&$86.64\pm0.32$&$85.65\pm0.72$&$87.43$\\
    % &balancedPU+ADT&$81.37\pm9.03$&$81.47\pm6.05$&$80.51\pm7.96$&$83.76\pm0.73$&$81.78$\\
    % &robustPU+ADT&$89.80\pm0.15$&$79.47\pm12.95$&$86.53\pm1.95$&$73.12\pm23.14$&$82.23$\\
    &BFGPU&\bm{$90.03\pm0.08$}&\bm{$88.51\pm1.05$}&\bm{$87.44\pm0.17$}&\bm{$86.50\pm0.40$}&\bm{$88.12$}\\
    \hline\hline
\end{tabular}
% \vskip -0.15in
\end{table*}
It is noticeable that when the imbalance ratio is high, the performance improvement brought by BFGPU is even greater compared to previous experiments. With larger values of $\sigma_{micro}$, we observed that many algorithms lack stability and may even fail. When algorithms fail, the evaluation metric F1 Score becomes highly unstable. This is because F1 Score does not treat positive and negative classes equally; more fundamentally, precision and recall are both centered around the positive class. Thus, the results in F1 Score can vary significantly depending on whether the classification leans towards the positive or negative class to the same extent. In contrast, the metric of average accuracy possesses greater stability and fairness.
\subsection{Experiments with Imbalance at Both Macro and Micro Levels}
On the SST-2 dataset, we set $\sigma_{macro}$ to 5 and 10 and explored how algorithm performance varied with $\sigma_{micro}$ set to [2, 4, 6, 8, 10].

Due to space limitations in the main text, the experimental statistical data on the SST-2 dataset are provided in \cref{sst25_table,sst210_table} in this appendix.
 \begin{table}[htbp]
\centering
\small
\caption{Experimental Results with Imbalance at Both Macro and Micro Levels when $\sigma_{macro=5}$.}
\label{sst25_table}
\begin{tabular}{c c c c c c c c}
\hline\hline
&\multirow{2}{*}{Method} &\multicolumn{5}{c}{AvgAcc} & \multirow{2}{*}{$AUC_{AvgAcc}$}\\\cline{3-7}
&&2&4&6&8&10\\
    \hline\hline
    \multirow{4}{*}{Macro}& Supervised&$72.65$&$70.46$&$69.32$&$63.52$&$55.42$&$66.27$\\
    % Coarse-Grained AD&$75.95\pm1.52$&$73.60\pm0.26$&$70.58\pm0.48$&$68.61\pm0.22$&$61.41\pm6.18$\\
    &DeepSAD&$51.93$&$54.79$&$52.17$&$45.91$&$49.22$&$50.80$\\
    &UnderSampling&$75.41$&$73.27$&$62.32$&$55.81$&$54.64$&$64.29$\\
    &OverSampling&$78.18$&$73.76$&$69.57$&$60.47$&$59.67$&$68.33$\\\hline
    \multirow{5}{*}{MIL}
    &MaxPooling&$50.00$&$50.00$&$50.00$&$50.00$&$57.87$&$51.57$\\
    &TopkPooling&$-$&$74.64$&$69.31$&$58.49$&$50.00$&$63.11$\\
    &Attention&$77.90$&$71.29$&$51.21$&$56.29$&$58.14$&$62.96$\\
    &FGSA&$50.00$&$50.00$&$50.00$&$50.00$&$50.00$&$50.00$\\
    &PReNET&$50.00$&$50.00$&$50.00$&$50.00$&$50.00$&$50.00$\\\hline
    \multirow{7}{*}{Micro}&DeepSAD&$50.37$&$53.13$&$49.52$&$45.60$&$50.39$&$49.80$\\
    &DeepSVDD&$51.38$&$54.29$&$50.00$&$48.74$&$53.10$&$51.50$\\
    &uPU&$71.45$&$77.89$&$57.25$&$50.00$&$50.00$&$61.32$\\
    &nnPU&$50.00$&$50.00$&$50.00$&$50.00$&$50.00$&$50.00$\\
    &balancedPU&$54.24$&$50.00$&$50.00$&$50.00$&$50.00$&$50.85$\\
    &robustPU&$75.60$&$76.57$&$58.21$&$50.00$&$50.00$&$62.08$\\
    % &uPU+ADT&$69.43\pm10.82$&$73.93\pm0.84$&$65.41\pm3.88$&$60.69\pm5.68$&$58.91\pm5.48$&$65.67$\\
    % &nnPU+ADT&$75.13\pm1.97$&$69.80\pm3.59$&$66.18\pm2.67$&$56.29\pm7.28$&$56.98\pm6.15$&$64.88$\\
    % &balancedPU+ADT&$69.98\pm2.76$&$64.19\pm11.47$&$48.07\pm2.39$&$51.89\pm7.35$&$50.39\pm1.45$&$56.90$\\
    % &robustPU+ADT&$79.01\pm7.05$&$72.44\pm9.16$&$66.91\pm12.77$&$58.49
    % \pm11.12$&$67.83\pm17.49$&$68.94$\\
    &BFGPU&\bm{$80.48$}&\bm{$78.22$}&\bm{$73.19$}&\bm{$63.84$}&\bm{$67.83$}&\bm{$72.71$}\\
    \hline\hline

&\multirow{2}{*}{Method} &\multicolumn{5}{c}{F1 Score} & \multirow{2}{*}{$AUC_{F1}$}\\\cline{3-7}
&&2&4&6&8&10&\\
    \hline\hline
    \multirow{4}{*}{Macro}& Supervised&$78.16$&$76.95$&$75.92$&$72.56$&$68.00$&$74.32$\\
    % Coarse-Grained AD&$75.54\pm1.54$&$73.11\pm0.25$&$70.16\pm0.61$&$68.05\pm0.13$&$60.93\pm6.09$\\
     &DeepSAD&$56.88$&$59.33$&$44.46$&$36.40$&$23.49$&$44.11$\\
     &UnderSampling&$69.83$&$75.00$&$71.28$&$69.84$&$56.86$&$68.56$\\
     &OverSampling&$70.56$&$65.79$&$62.37$&$60.47$&$62.90$&$64.42$\\\hline
    \multirow{5}{*}{MIL}
    &MaxPooling&$66.67$&$66.67$&$66.67$&$66.67$&$66.67$&$66.67$\\
    &TopkPooling&$-$&$75.00$&$69.38$&$68.00$&$66.67$&$69.76$\\
    &Attention&$81.46$&$77.75$&$67.21$&\bm{$69.70$}&$70.73$&$73.37$\\
    
    &FGSA&$66.67$&$66.67$&$66.67$&$66.67$&$66.67$&$66.67$\\
    &PReNET&$66.67$&$66.67$&$66.67$&$66.67$&$66.67$&$66.67$\\\hline
    \multirow{7}{*}{Micro}&DeepSAD&$47.48$&$51.85$&$44.26$&$38.50$&$47.01$&$45.82$\\
    &DeepSVDD&$48.99$&$51.81$&$44.85$&$48.33$&$50.36$&$48.87$\\
    &uPU&$77.09$&$78.81$&$68.25$&$66.67$&$66.67$&$71.50$\\
    &nnPU&$66.67$&$66.67$&$66.67$&$66.67$&$66.67$&$66.67$\\
    &balancedPU&$13.85$&$0.00$&$0.00$&$0.00$&$0.00$&$2.77$\\
    &robustPU&$80.07$&$80.74$&$71.03$&$66.67$&$66.67$&$73.04$\\
    % &uPU+ADT&$59.27\pm30.49$&$78.52\pm0.74$&$51.64\pm25.64$&$45.00\pm25.61$&$42.52\pm23.95$&$55.39$\\
    % &nnPU+ADT&$79.35\pm1.25$&$73.85\pm3.35$&$73.71\pm1.13$&$59.17\pm15.42$&$45.58\pm39.51$&$66.33$\\
    % &balancedPU+ADT&$74.73\pm1.19$&$61.39\pm21.25$&$39.93\pm18.26$&$50.93\pm3.70$&$59.91\pm5.12$&$57.38$\\
    % &robustPU+ADT&$79.32\pm9.41$&$70.77\pm18.28$&$63.41\pm22.24$&$47.67\pm41.56$&$73.50\pm10.40$&$66.93$\\
    &BFGPU&\bm{$83.48$}&\bm{$81.59$}&\bm{$78.76$}&$64.92$&$\bm{74.24}$&\bm{$76.60$}\\
    \hline\hline
\end{tabular}
% \vskip 0.2in
\end{table}
\begin{table}[ht]
\centering
\small
\caption{Experimental Results with Imbalance at Both Macro and Micro Levels when $\sigma_{macro=10}$.}
\label{sst210_table}
\begin{tabular}{c c c c c c c c}
\hline\hline
&\multirow{2}{*}{Method} &\multicolumn{5}{c}{AvgAcc} & \multirow{2}{*}{$AUC_{AvgAcc}$}\\\cline{3-7}
&&2&4&6&8&10\\
    \hline\hline
    \multirow{4}{*}{Macro}& Supervised&68.78	&54.95	&50.00	&50.00	&50.00	&54.75\\
    &DeepSAD&66.02	&56.93	&53.62	&39.06	&55.81	&54.29\\
    &UnderSampling&77.62	&70.79	&60.14	&50.00	&53.48	&62.41\\
    &OverSampling&69.89	&66.83	&61.59	&62.26	&54.65	&63.04\\\hline
    \multirow{5}{*}{MIL}
    &MaxPooling&$50.00$&$50.00$&$50.00$&$50.00$&$50.00$&$50.00$\\
    &TopkPooling&$-$&72.77	&50.72	&50.00	&50.00	&55.87\\
    &Attention&75.41	&50.99	&71.01	&50.00	&50.00&	59.48\\
    &FGSA&$50.00$&$50.00$&$50.00$&$50.00$&$50.00$&$50.00$\\
    &PReNET&77.62	&72.77	&65.94	&74.52	&50.00	&68.17\\\hline
    \multirow{7}{*}{Micro}&DeepSAD&47.51	&51.49	&43.48	&49.06	&47.67&	47.84\\
    &DeepSVDD&56.60	&58.49	&51.16	&58.14	&45.35	&53.95\\
    &uPU&74.03	&66.34	&50.00	&50.00	&50.00	&58.07\\
    &nnPU&$50.00$&$50.00$&$50.00$&$50.00$&$50.00$&$50.00$\\
    &balancedPU&$50.00$&$50.00$&$50.00$&$50.00$&$50.00$&$50.00$\\
    &robustPU&74.86	&63.37	&50.00	&50.00	&50.00	&57.65\\
    &BFGPU&\bm{$79.01$}&\bm{$78.22$}&\bm{$78.99$}&\bm{$76.42$}&\bm{$68.60$}&\bm{$76.25$}\\
    
    \hline\hline
&\multirow{2}{*}{Method} &\multicolumn{5}{c}{F1 Score} & \multirow{2}{*}{$AUC_{F1}$}\\\cline{3-7}
&&2&4&6&8&10&\\
    \hline\hline
    \multirow{4}{*}{Macro}& Supervised&75.80&68.51&66.67&66.67&66.67&68.86\\
     &DeepSAD&74.43&65.32&53.62&56.88&32.14&56.48\\
     &UnderSampling&77.56&63.80&49.54&66.67&23.08&56.13\\
     &OverSampling&78.87&70.73&74.36&76.19&66.67&73.36\\
\hline
    \multirow{5}{*}{MIL}
    &MaxPooling&0.00&0.00&0.00&0.00&0.00&0.00\\
    &TopkPooling&$-$&78.43&66.99&66.67&66.67&69.69\\
    &Attention&79.91&67.11&77.27&66.67&66.67&71.53
\\
    
    &FGSA&$66.67$&$66.67$&$66.67$&$66.67$&$66.67$&$66.67$\\
    &PReNET&81.29&79.39&78.26&74.59&66.67&76.04\\\hline
    \multirow{7}{*}{Micro}&DeepSAD&53.66&59.17&50.63&51.79&60.18&55.09\\
    &DeepSVDD&64.26&44.75&59.63&63.83&67.19&59.93\\
    &uPU&83.56&71.67&$66.67$&$66.67$&$66.67$&71.05\\
    &nnPU&$66.67$&$66.67$&$66.67$&$66.67$&$66.67$&$66.67$\\
    &balancedPU&0.00&0.00&0.00&0.00&0.00&0.00\\
    &robustPU&79.55&73.19&66.67&66.67&66.67&70.55\\
    &BFGPU&\bm{$87.27$}&\bm{$81.48$}&\bm{$78.52$}&\bm{$76.19$}&\bm{$67.19$}&\bm{$78.13$}\\
    \hline\hline
\end{tabular}
% \vskip 0.2in
\end{table}
In settings where both imbalances coexist, with the total proportion of negative information in the text dataset reaching its extreme low, we found that most comparative methods failed. However, BFGPU still maintained stable and excellent performance in this extreme scenario, demonstrating its strong versatility.
\subsection{Ablation Study}
To explore the necessity of each component of BFGPU, we conducted ablation experiments, comparing the algorithm's performance without the new loss function, without pseudo-labeling, and without the threshold adjustment technique. We performed these experiments on the IMDB and SST-2 datasets, setting $\sigma_{micro}$ to [2, 4, 6, 8, 10]. Due to space limitations in the main text, the additional experiments on the SST-2 dataset are provided in \cref{ablation_table1} in this appendix.
\begin{table}[htbp]
\centering
\small
\caption{The table presents AvgAcc and F1 results of the ablation study under varying values of $\sigma_{\text{micro}} \in \{2, 3, 4, 5\}$ on the IMDB dataset.}
\label{ablation_table2}
\begin{tabular}{c c c c c c c}
\hline\hline
% \multirow{2}{*}{Decoupling}
\multirow{2}{*}{PU Loss}&\multirow{2}{*}{Pseudo Labels}&\multirow{2}{*}{Threshold}&\multicolumn{4}{c}{AvgAcc} \\\cline{4-7}&&&$2$&$3$&$4$&$5$\\
% & \multirow{2}{*}{$AUC_{AvgAcc}$}
\hline\hline
$\times$&$\checkmark$&$\checkmark$&$84.98 \pm 3.14$&$84.27 \pm 2.12$&$85.55 \pm 0.31$&$62.19\pm10.22$\\
% &$81.00$
$\checkmark$&$\times$&$\checkmark$&$88.13\pm0.76$&$86.04\pm0.66$&$85.40\pm0.18$&$83.02\pm1.35$\\
% &$78.47$
$\checkmark$&$\checkmark$&$\times$&\bm{$88.25\pm0.90$}&$86.97\pm0.89$&$85.40\pm0.18$&$81.18\pm2.13$\\
% &$70.55$
$\checkmark$&$\checkmark$&$\checkmark$&$88.13\pm1.07$&\bm{$87.93\pm0.58$}&\bm{$86.02\pm0.69$}&\bm{$83.64\pm0.78$}\\
% &\bm{$83.03$}
\hline\hline
\multirow{2}{*}{PU Loss}&\multirow{2}{*}{Pseudo Labels}&\multirow{2}{*}{Threshold}&\multicolumn{4}{c}{F1 Score}\\\cline{4-7}&&&$2$&$3$&$4$&$5$\\
 % & \multirow{2}{*}{$AUC_{F1}$}
\hline\hline
$\times$&$\checkmark$&$\checkmark$&$82.34\pm2.83$&$84.06\pm2.37$&$85.67\pm0.43$&$58.35\pm13.32$\\
% &$83.31$
$\checkmark$&$\times$&$\checkmark$&\bm{$88.63\pm0.55$}&$86.95\pm0.42$&$85.01\pm0.88$&$84.36\pm0.74$\\
% &$74.20$
$\checkmark$&$\checkmark$&$\times$&$88.44\pm0.97$&$86.50\pm1.12$&$85.01\pm0.88$&$78.92\pm3.39$\\
% &$65.08$
$\checkmark$&$\checkmark$&$\checkmark$&$88.12\pm1.24$&\bm{$88.36\pm0.58$}&\bm{$86.57\pm0.69$}&\bm{$84.61\pm0.43$}\\
% &\bm{$84.64$}
\hline\hline
\end{tabular}
\end{table}
\begin{table}[htbp]
\centering
\small
\caption{The table presents AvgAcc and F1 results of the ablation study under varying values of $\sigma_{\text{micro}} \in \{2,  4, 6,8,10\}$ on the SST-2 dataset.}
\label{ablation_table1}
\resizebox{1\textwidth}{!}{\begin{tabular}{c c c c c c c c}
\hline\hline
% \multirow{2}{*}{Decoupling}
\multirow{2}{*}{PU Loss}&\multirow{2}{*}{Pseudo Labels}&\multirow{2}{*}{Threshold}&\multicolumn{5}{c}{AvgAcc} \\\cline{4-8}&&&$2$&$4$&$6$&$8$&$10$\\
% & \multirow{2}{*}{$AUC_{AvgAcc}$}
\hline\hline
$\times$&$\checkmark$&$\checkmark$&$85.27\pm1.50$&$83.33\pm0.84$&$79.95\pm2.46$&$77.36\pm2.67$&$79.07\pm3.80$\\
% &$81.00$
$\checkmark$&$\times$&$\checkmark$&$87.75\pm2.08$&\bm{$86.14\pm1.07$}&$71.50\pm14.69$&$66.35\pm12.76$&$80.62\pm3.33$\\
% &$78.47$
$\checkmark$&$\checkmark$&$\times$&$86.83\pm0.13$&$72.11\pm12.07$&$74.15\pm1.49$&$58.81\pm5.78$&$60.85\pm7.91$\\
% &$70.55$
$\checkmark$&$\checkmark$&$\checkmark$&\bm{$88.40\pm0.68$}&$82.51\pm0.62$&\bm{$82.13\pm0.90$}&\bm{$79.56\pm1.60$}&\bm{$82.56\pm1.64$}\\
% &\bm{$83.03$}
\hline\hline
\multirow{2}{*}{PU Loss}&\multirow{2}{*}{Pseudo Labels}&\multirow{2}{*}{Threshold}&\multicolumn{5}{c}{F1 Score}\\\cline{4-8}&&&$2$&$4$&$6$&$8$&$10$\\
 % & \multirow{2}{*}{$AUC_{F1}$}
\hline\hline
$\times$&$\checkmark$&$\checkmark$&$86.65\pm0.92$&$84.95\pm0.64$&$82.74\pm1.33$&$80.87\pm2.12$&$81.32\pm2.96$\\
% &$83.31$
$\checkmark$&$\times$&$\checkmark$&$88.41\pm1.80$&\bm{$87.31\pm0.95$}&$61.39\pm31.32$&$51.26\pm36.66$&$82.65\pm2.92$\\
% &$74.20$
$\checkmark$&$\checkmark$&$\times$&$86.88\pm0.54$&$71.80\pm10.90$&$68.84\pm2.64$&$40.53\pm8.76$&$57.33\pm12.78$\\
% &$65.08$
$\checkmark$&$\checkmark$&$\checkmark$&\bm{$88.73\pm0.61$}&$84.13\pm0.30$&\bm{$83.96\pm1.10$}&\bm{$81.95\pm1.87$}&\bm{$84.41\pm1.64$}\\
% &\bm{$84.64$}
\hline\hline
\end{tabular}}
\end{table}
The experimental results indicate that each component of BFGPU is crucial, collectively ensuring the algorithm's stability and excellent performance.
\subsection{Experiments with Different Base Model}
To verify that our results are not limited by the choice of backbone model, we conducted further evaluations using DeBERTa. The results are shown in \cref{Deberta}. 
\begin{table}[htbp]
\centering
\small
\caption{The table presents AvgAcc and F1 scores of various algorithms with the Deberta model under varying values of $\sigma_{\text{micro}} \in \{2,  4, 6,8, 10\}$, along with the estimated AUC with respect to changes in AvgAcc and F1 on the SST-2 Dataset.}
\label{Deberta}
% \vskip 0.15in
\begin{tabular}{c c c c c c c c}
\hline\hline
&\multirow{2}{*}{Method} &\multicolumn{5}{c}{AvgAcc} &\multirow{2}{*}{$AUC_{AvgAcc}$}\\\cline{3-7}&&2&4&6&8&10&\\
    \hline\hline
\multirow{2}{*}{Macro}& Supervised&	81.22&	77.72&	78.26&	77.36&	77.91&78.49\\
&DeepSAD&	49.17&	59.41&	42.75&	50.94&	47.67&49.99\\\hline
\multirow{5}{*}{MIL}& MaxPooling&	75.97&	70.30&	76.81&	65.09&	50.00&67.63\\
&TopkPooling&	$-$&	74.26&	77.54&	78.30&	73.26&75.84\\
&Attention&	72.65&	65.35&	52.90&	50.83&	50.94&58.53\\
&FGSA&	73.48&	64.36&	54.35&	50.94&	50.00&58.63\\
&PReNET&	71.82&	61.39&	65.22&	55.66&	61.63&63.14\\\hline
\multirow{7}{*}{Micro}& DeepSAD&	48.34&	49.51&	51.45&	51.89&	52.33&50.70\\
&DeepSVDD&	48.07&	40.10&	52.90&	51.89&	51.16&48.82\\
&uPU&	72.65&	57.43&	72.46&	59.43&	67.44&65.88\\
&nnPU&	72.65&	50.50&	50.00&	50.00&	50.00&54.63\\
&balancedPU&	82.32&	80.20&	78.26&	76.42&	72.93&78.03\\
&robustPU&	83.98&	80.20&	76.47&	83.33&	80.19&80.50\\
&BFGPU&	\bm{$88.40$}&	\bm{$81.68$}&	\bm{$78.99$}&	\bm{$83.96$}&	\bm{$81.40$}&\bm{$82.89$}\\\hline\hline
&\multirow{2}{*}{Method} &\multicolumn{5}{c}{F1 Score} &\multirow{2}{*}{$AUC_{F1}$}\\\cline{3-7}&&2&4&6&8&10&\\
    \hline\hline
\multirow{2}{*}{Macro}&Supervised&	80.79&	77.61&	79.73&	76.92&	79.12&78.83\\
&DeepSAD&	55.34&	69.40&	56.83&	60.61&	62.81&61.00\\\hline
\multirow{5}{*}{MIL}& MaxPooling&	65.37&	63.42&	59.76&	50.04&	53.70&58.46\\
&TopkPooling&	$-$&	63.23&	64.87&	53.44&	55.46&59.00\\
&FGSA&	65.71	&55.56	&22.22&	3.70&	0.00&29.44\\
&Attention	&63.47	&47.76	&35.29	&23.08	&10.71&36.06\\
&PReNET	&74.75	&40.00	&47.83	&25.40	&37.74&45.14\\\hline
\multirow{7}{*}{Micro}&DeepSAD&	63.55&	66.23&	57.32&	61.65&	67.72&63.29\\
&DeepSVDD	&64.26&	44.75&	59.63&	63.83&	67.19&59.93\\
&uPU	&64.52&	33.85&	66.67&	41.10&	62.16&53.66\\
&nnPU	&78.15&	66.89&	66.67&	66.67&	66.67&69.01\\
&balancedPU	&84.08&	82.91&	81.71&	79.67&	81.19&81.91\\
&robustPU	&82.53&	77.78&	80.77&	82.44&	75.32&79.77\\
&BFGPU	&\bm{$88.65$}&	\bm{$84.12$}&	\bm{$81.76$}&	\bm{$85.96$}&	\bm{$83.33$}&\bm{$84.76$}\\\hline\hline
\end{tabular}
\end{table}
\subsection{Experiments with Different PU Loss with Pseudo Labeling and ADT}
o further demonstrate the superiority of
the BFGPU loss function, we compared its performance against other PU learning methods when
integrated with our proposed pseudo-labeling and ADT strategies. The results are shown in \cref{ADT SST2}.
\begin{table}[htbp]
\centering
\small
\caption{The table presents the performance of various PU learning methods when integrated with our proposed pseudo-labeling and ADT strategies. }
\label{ADT SST2}
\begin{tabular}{c c c c c c c}
\hline\hline
\multirow{2}{*}{Method} &\multicolumn{5}{c}{AvgAcc} & \multirow{2}{*}{$AUC_{AvgAcc}$}\\\cline{2-6}
&2&4&6&8&10\\
\hline\hline
uPU&$82.69\pm2.02$&$76.07\pm1.82$&$71.26\pm3.36$&$64.47\pm0.89$&$61.24\pm4.78$&$71.15$\\
nnPU&$87.02\pm0.60$&\bm{$83.83\pm1.42$}&$79.23\pm1.49$&$75.16\pm2.48$&$74.03\pm3.95$&$79.85$\\
balancedPU&$79.83\pm6.66$&$77.23\pm5.73$&$68.84\pm11.79$&$66.04\pm5.82$&$72.09\pm1.64$&$72.81$\\
robustPU&$86.00\pm1.84$&$74.09\pm20.01$&$78.02\pm2.33$&$72.96\pm8.02$&$61.63\pm14.52$&$74.54$\\
BFGPU&\bm{$88.40\pm0.68$}&$82.51\pm0.62$&\bm{$82.13\pm0.90$}&\bm{$79.56\pm1.60$}&\bm{$82.56\pm1.64$}&\bm{$83.03$}\\
\hline\hline
\multirow{2}{*}{Method} &\multicolumn{5}{c}{F1 Scroe} & \multirow{2}{*}{$AUC_{F1}$}\\\cline{2-6}
&2&4&6&8&10\\
    \hline\hline

uPU&$84.09\pm1.08$&$77.99\pm1.55$&$75.59\pm2.66$&$69.91\pm1.50$&$65.42\pm9.19$&$74.60$\\
    nnPU&$87.75\pm0.60$&\bm{$85.13\pm0.84$}&$81.56\pm1.00$&$78.85\pm2.79$&$78.39\pm2.57$&$82.34$\\
    balancedPU&$80.94\pm6.40$&$79.54\pm4.81$&$57.02\pm32.67$&$65.03\pm13.16$&$76.11\pm1.94$&$71.73$\\
    robustPU&$86.91\pm1.48$&$59.86\pm45.37$&$80.23\pm2.51$&$74.05\pm8.97$&$48.39\pm42.41$&69.89\\
    BFGPU&\bm{$88.73\pm0.61$}&$84.13\pm0.30$&\bm{$83.96\pm1.10$}&\bm{$81.95\pm1.87$}&\bm{$84.41\pm1.64$}&\bm{$84.64$}\\
    \hline\hline
\end{tabular}
\end{table}

\subsection{Sensitive Analysis}
We have supplemented the sensitivity analysis of hyperparameters $\lambda_{bfgpu}$ and $\lambda_{pse}$. We completed experiments on the IMDB dataset, setting $\sigma_{micro} = 5$. We set the variation range of the two parameters to [1, 2, 3, 4, 5], and kept one constant while varying the other. Due to space limitations in the main text, the sensitive analysis of $\lambda_{pse}$ on the SST-2 dataset are provided in \cref{sen2} in this appendix.
\begin{table}[htbp]
    \centering
    \captionof{table}{This table presents the AvgAcc of the sensitive analysis of $\lambda_{bfgpu}$ on the SST-2 dataset.}  
    \label{sen1}
    \begin{tabular}{c c c c c c}
    \hline\hline
    $\lambda_{bfgpu}$&1&2&3&4&5\\\hline\hline
    $AvgAcc$&83.41&84.72&84.99&84.55&82.79\\
    $F1 Score$&84.56&85.68&85.83&85.56&84.45\\
    \hline\hline
    \end{tabular}
\end{table}
\begin{table}[htbp]
\centering
\captionof{table}{This table presents the AvgAcc of the sensitive analysis of $\lambda_{pse}$ on the SST-2 dataset.}  
\label{sen2}
\begin{tabular}{c c c c c c}
\hline\hline
$\lambda_{pse}$&1&2&3&4&5\\\hline\hline
$AvgAcc$&83.41&84.29&84.68&84.64&84.94\\
$F1 Score$&84.56&85.23&85.49&85.51&85.67\\
\hline\hline
\end{tabular}
\end{table}
Experiments have demonstrated that the performance of the algorithm remains relatively stable under different hyperparameter settings, so there is no need to worry too much about hyperparameter tuning.

\end{document}